\pdfoutput=1

\documentclass[11pt]{article}

\usepackage[final]{acl2024}
\usepackage{float}

\usepackage{xcolor}

\usepackage{color,soul}

\usepackage{times}
\usepackage{latexsym}
\usepackage{enumitem}

\usepackage[T1]{fontenc}

\usepackage[utf8]{inputenc}

\usepackage{microtype}

\usepackage{inconsolata}
\usepackage{svg}
\usepackage{authblk}

\usepackage{tabularx}
\usepackage{booktabs}
\usepackage{times}
\usepackage{latexsym}
\usepackage{subcaption}
\usepackage{graphicx}
\usepackage{amsmath, amsthm, amssymb}
\usepackage[T1]{fontenc}

\usepackage{multicol}

\usepackage[utf8]{inputenc}

\usepackage{graphicx}
\usepackage{array}

\title{Few shot chain-of-thought driven reasoning to prompt LLMs for open-ended medical question answering}

\newcommand{\equalcontrib}{\textsuperscript{*}}

\author[2]{Saeel Sandeep Nachane\thanks{Equal contribution.}}
\author[1]{Ojas Gramopadhye\equalcontrib}
\author[2]{Prateek Chanda\equalcontrib}
\author[2]{Ganesh Ramakrishnan}
\author[2]{\\Kshitij Sharad Jadhav}
\author[1]{Yatin Nandwani}
\author[1]{Dinesh Raghu}
\author[1]{Sachindra Joshi}

\affil[1]{IBM Research}
\affil[2]{Indian Institute of Technology Bombay}

\affil[1]{\texttt{\{ojas.gramopadhye@, yatin.nandwani@, diraghu1@in, jsachind@in\}}.ibm.com}
\affil[2]{\texttt{\{20d180028@, prateekch@cse ,ganesh@cse, kshitij.jadhav@\}}.iitb.ac.in}

\begin{document}

\newcommand{\fivepointscale}{\colorbox{teal}{Strongly Agree}, Agree, Neutral, Disagree, \colorbox{magenta}{Strongly Disagree}}
\newcommand{\threepointscale}{Agree, Neutral, \colorbox{red}{Disagree}}
\newcommand{\ClinicianCases}{\textsc{ClinicianCases}}

\newcommand{\MedQAOpen}{\textsc{MedQA-Open}}
\newcommand{\MedQAOriginal}{\textsc{MedQA-MCQ}}
\newcommand{\MCQEliminative}{\textsc{MCQ-Eliminative}}
\newcommand{\OpenEliminative}{\textsc{Eliminative}}
\newcommand{\MCQClinicR}{\textsc{MCQ-ClinicR}}
\newcommand{\OpenClinicR}{\textsc{ClinicR}}

\maketitle
\begin{abstract}
In this paper, we propose a modified version of the \MedQAOriginal\ dataset, named \MedQAOpen{}, which contains open-ended medical questions without options to mimic clinical scenarios, along with clinician-approved reasoned answers. Additionally, we implement a prompt driven by Chain of Thought (CoT) reasoning, \OpenClinicR{}, to mirror the prospective process of incremental reasoning, reaching a correct response to medical questions. We empirically demonstrate how \OpenClinicR{} outperforms the state-of-the-art 5-shot CoT-based prompt \cite{lievin2022can}. We also present an approach that mirrors real-life clinical practice by first exploring multiple differential diagnoses through \MCQClinicR{}, and subsequently narrowing down to a final diagnosis using \MCQEliminative{}. Finally, emphasizing the importance of response verification in medical settings, we utilize a reward model mechanism, replacing the elimination process performed by \MCQEliminative{}.\footnote{The code and prompts used in our experiments are available  at \url{https://github.com/ColdSeal/ClinicR} }


\end{abstract}

\section{Introduction}

Large Language models (LLMs) are increasingly utilized in the healthcare sector, particularly for patient query-related tasks. These LLM-driven tools could potentially interpret and respond to patient inquiries, and provide information on symptoms, diseases, treatments, and healthcare guidelines \cite{thirunavukarasu2023large}. By analyzing vast amounts of medical literature and data, LLMs could also offer precise, up-to-date responses, improving patient education and engagement \cite{singhal2022large}. The ability of LLMs to process natural language queries makes them accessible and user-friendly, thus enhancing the patient experience and satisfaction \cite{clusmann2023future}. As technology evolves, LLMs are expected to play a pivotal role in delivering personalized healthcare information, contributing to informed decision-making and better health outcomes \cite{clusmann2023future}. However, with the recent advances in prompt engineering techniques in the Large Language Models space, there is an underlying requirement for accuracy and verifiability. 

\begin{figure*}[t!]
    \centering
        \includegraphics[width=\linewidth]{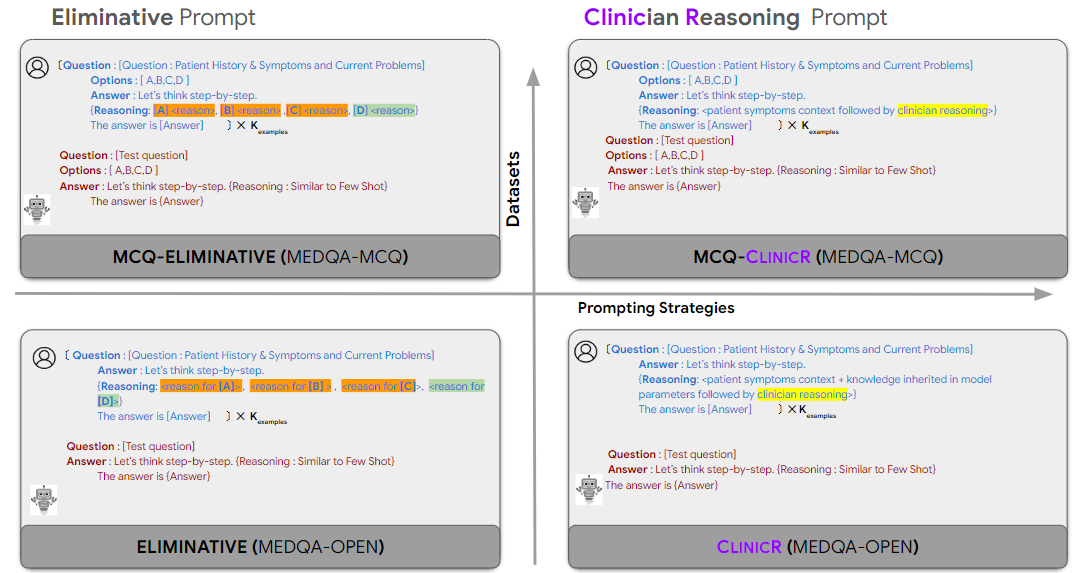}
    \caption{An overview of the two prompting strategies used for predicting the answer for both the MCQ version (\MedQAOriginal{}) and open-ended version (\MedQAOpen{}) of the MedQA dataset.}
    \label{fig:promptvsDatasets}
\end{figure*}

\subsection{Key Contributions}


We propose techniques to prompt large language models to generate answers to healthcare-related questions. We build upon the MedQA-USMLE dataset with the help of medical experts and construct a novel medical question corpus that contains open-ended medical questions, named \MedQAOpen, along with expert-verified appropriate reasoning along with the answer.

We propose the \OpenClinicR\ designed to predict an answer to the medical questions by incremental reasoning. This 5-shot prompt (Appendix A) uses Chain-of-Thought reasoning drawing successive inferences from a given patient history. This prompt integrates medical findings, and investigation reports to predict the final response. In our comparisons, we show that the \OpenClinicR\ outperforms the 5-shot, CoT-prompt proposed by~\citet{lievin2022can}. We demonstrate that prospective, incremental and reasoning-driven prompting, that mimics real-life clinical scenarios, performs significantly better at answering open-ended medical questions.

Further, we implemented a method of developing several differential diagnoses through \OpenClinicR\ followed by trimming down to a final diagnosis using \MCQEliminative\ mirroring real-life clinical practice. This \textit{forward-backward} method mimics real-life clinicians' thought processes.  We created a reward modelling dataset for building a verifier on medical question-answer pairs. In this dataset, we include {\em Correct} and {\em Incorrect} Question-Reasoning-Answer triplets, a representative sample of which have been evaluated and verified by medical experts. We use the \MedQAOriginal{} dataset for the Question-Answer pair and artificially generate the Reasoning. We also release the reward model \textit{verifier} trained with this dataset. Finally, we use the reward model in place of the \MCQEliminative\ prompt to trim down the available options in the \textit{forward-backward} method to generate the final diagnosis.
\section{Background \& Related Work}
The USMLE-MedQA dataset \citep{jin2021disease}, a medical exam dataset, consists of questions sourced from professional medical board exams in the USA. The dataset is a comprehensive resource for investigating medical question answering in a standardized MCQ examination format. The dataset offers complexity and challenge, however, the current version of the dataset is not practically useful to be deployed in a patient query-based healthcare system since four options are associated with each question thereby not being similar to in-clinic scenarios.

Recent advancements in the application of Large Language Models (LLMs) for medical question answering have highlighted the potential of these models to demonstrate applicability in medical diagnostics, education, and research. \cite{wang2023augmenting} explored augmenting black-box LLMs with medical textbooks to enhance clinical question answering. Similarly, \cite{yagnik2024medlm} introduced MedLM, exploring the efficacy of LLMs in handling diverse medical question-answering systems. 
Furthermore, recent research highlights prompt engineering as a viable alternative to traditional fine-tuning approaches. \cite{maharjan2024openmedlm} demonstrated that prompt-based methods outperform fine-tuning when working with open-source medical language models, particularly in resource-constrained environments. Another noteworthy contribution is by \cite{Shi2023MKRAGMK}, who proposed Mededit, a framework for model editing using external knowledge bases to improve performance in medical QA. Lastly, \cite{sayin2024can} examined effective interaction methods between LLMs and medical professionals, questioning whether LLMs can correct physicians' responses, thus opening new avenues for improving collaboration in medical domains.
The exploration of LLMs in generating not only accurate but also reasoning-based responses to medical questions, marks a significant step forward. Models such as PubMedGPT~\cite{ bolton2022pubmedgpt} and Codex~\citep{lievin2022can} have established benchmarks on datasets like MedQA through innovative approaches, including classification head, Chain-of-Thought, and knowledge grounding, highlighting the importance of not just what is answered, but how the answer is derived. These studies lay the groundwork for further advancements in medical QA using LLMs, particularly in developing more efficient and transparent interaction strategies to enhance clinical outcomes.

Answering medical questions involves the requirement of fact verification as well as the necessity of reasoning skills. These have been attempted to be addressed through the following methods:

\vspace{0.5ex}
\noindent {\bf Fact verification in LLMs:} Fact verification in LLMs involves computational linguistics, AI, and digital media. \citet{guo2022survey} present detailed approaches to automated fact-checking, including claim detection, evidence retrieval, and verification. This process assesses truthfulness, identifies evidence, and generates explanations, enhancing robust verification systems~\citep{guo2022survey}.

\vspace{0.5ex}
\noindent {\bf Chain-of-Thought Prompting:} Initially, scaling language models up appeared to benefit more knowledge-intensive tasks than reasoning-heavy ones \citep{rae2022scaling}. Nevertheless, \citet{wei2022chain} demonstrated that LLMs could be applied to reasoning-intensive problems by prompting the model to generate step-by-step solutions, coined “Chain-of-Thought” (CoT). CoT prompting led to substantial improvements in many reasoning-intensive tasks~\cite{wei2022chain,zhou2022learning,drozdov2022compositional,nye2021show}, allowing to bridge the gap with human-level performances.  The work of \citet{hendrycks2020measuring} evaluates LLMs using the MMLU dataset across diverse subjects like medicine, showcasing their broad applicability. Additionally, \citet{mesinovic2023explainable} emphasize the importance of model interpretability and explainability for trust in medical applications.

\begin{table*}[t!]
\centering
\begin{tabular}{p{0.5\linewidth} | p{0.4\linewidth}}
\toprule
\textbf{\MedQAOriginal} & \textbf{\MedQAOpen} \\
\midrule
\small{\textbf{\textcolor{blue}{Question}}: Four weeks after starting hydrochlorothiazide, a 49-year-old man with hypertension comes to the physician because of muscle cramps and weakness. His home medications also include amlodipine. His blood pressure today is 176/87 mm Hg. Physical examination shows no abnormalities. The precordial leads of a 12-lead ECG are shown.

The addition of \textbf{\hl{which of the following}} is most likely to have prevented this patient's condition? 

\textbf{(A)} Torsemide \textbf{(B)} Nifedipine \textbf{(C)} Eplerenone \textbf{(D)} Hydralazine
} & 
\small{\textbf{\textcolor{blue}{Question}}: Four weeks after starting hydrochlorothiazide, a 49-year-old man with hypertension comes to the physician because of muscle cramps and weakness. His home medications also include amlodipine. His blood pressure today is 176/87 mm Hg. Physical examination shows no abnormalities. The precordial leads of a 12-lead ECG are shown. The addition of  \textbf{\hl{which medication is most likely}} to have prevented this patient's condition?}

\\
\bottomrule
\end{tabular}
\caption{\label{citation-guide}
Sample question from the \MedQAOriginal{} dataset and its counterpart in \MedQAOpen{} dataset. For more details, check Appendix B
}
\label{tab:example-converted-data}
\end{table*}

\begin{figure*}[t!]
    \centering
        \includegraphics[width=\linewidth]{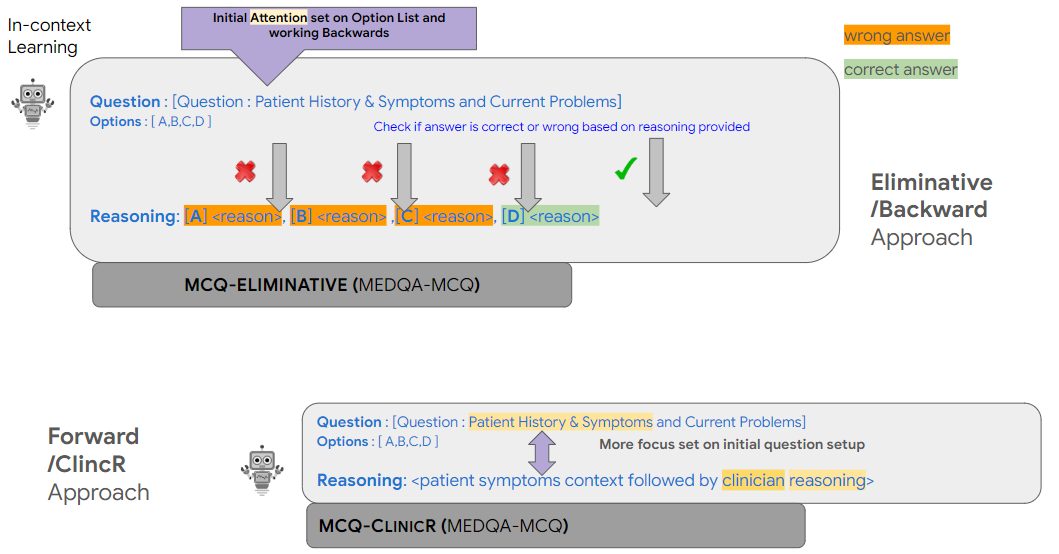}
    \caption{\MCQEliminative{} contains an eliminative form of reasoning, which iterates over options, accepting or discarding them as per their correctness. Often this does not cater to the context of real-life clinical investigation, unlike the incremental reasoning method as per \MCQClinicR{}.}
    \label{fig:Eliminative_vs_Forward_Motiv}
\end{figure*}

\begin{figure*}[t!]
    \centering
    \begin{subfigure}[t]{\textwidth}
        \centering
\includegraphics[width=\linewidth]{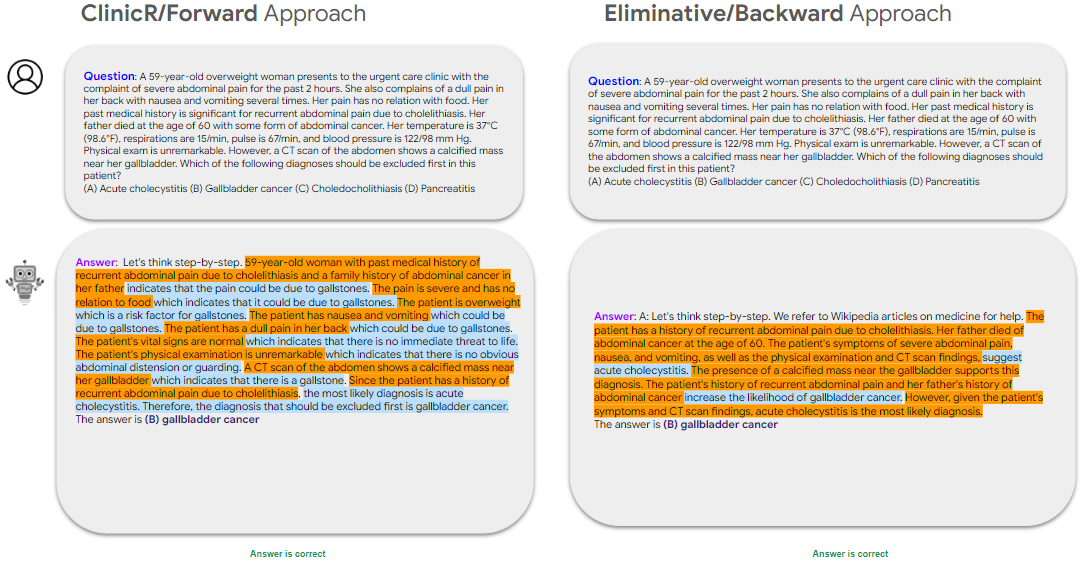}
        \caption{Responses produced by \MCQEliminative{} vs \MCQClinicR{} on the \MedQAOriginal{} dataset}
    \end{subfigure}%

    \begin{subfigure}[t]{\textwidth}
        \centering
\includegraphics[width=\linewidth]{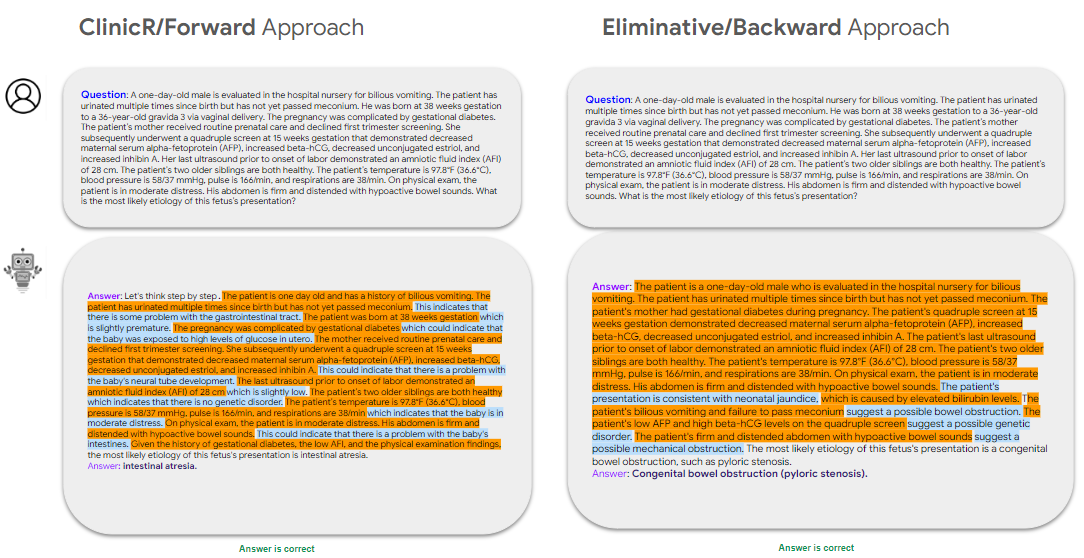}
        \caption{Responses produced by \OpenEliminative{} vs \OpenClinicR{} on \MedQAOpen{} dataset}
    \end{subfigure}
    \caption{Illustrative Example: Showcasing two different Prompting Strategies with their responses across two dataset variants: (a) \MedQAOriginal{} and (b) \MedQAOpen. \colorbox{orange}{context of the answer} and \colorbox{cyan}{corresponding reasoning} are highlighted.}
    \label{fig: codex_vs_MedCodex_responses}
\end{figure*}

\section{Approach}

In this section, we first define a task which can be used to evaluate the capability of large language models (LLMs) to be used as clinical support systems. We then describe prompt-based and verifier-based approaches to solve the task.

\subsection{Task Definition}
The primary goal of this work is to evaluate the capabilities of LLMs in answering questions for real-world clinical decision support systems. Such a realistic setup requires high-quality medical reasoning over a complex medical scenario. The MedQA dataset~\citep{zhang2018medical} is a publicly available collection of complex medical questions with multiple choices based on the
United States medical license exams. To emulate real-world medical scenarios, we convert these multiple-choice questions into open-ended questions by (1) removing the multiple-choice options and (2) re-pharsing the question to be open-ended. An example of an MCQ question and its open-ended counterpart is shown in Table \ref{tab:example-converted-data}. We refer to the original MCQ-based dataset as \MedQAOriginal{}, and the open-ended counterpart as \MedQAOpen{}.

In addition to \MedQAOpen{} dataset, we also created 25 real-life complex medical Question-Answer pairs from practising medical professionals, based on cases encountered by them. We refer to this as the \ClinicianCases{} dataset.

\subsection{Eliminative Prompt}
\citet{lievin2022can} devised a few-shot Chain-of-Thought (CoT) prompt for the \MedQAOriginal\ dataset. The prompt triggers the model to first copy relevant patient information required to answer the multiple-choice question, and then each option is considered in the order they appear in the question. During the consideration of each option, the model is expected to provide reasoning for selecting or rejecting it. Once a correct option is encountered, the model stops the CoT reasoning and generates the final answer. We refer to this prompt as \MCQEliminative\ prompt.

We adapt this prompt for the open setting by making the following changes: (1) replacing the multiple-choice question with its corresponding open-ended question, (2) removing explicit mentions of options in the reasoning, and (3) replacing any references made using the option index with the text of that option. For example, \textit{The answer is (D)} is replaced with \textit{The answer is Eplerenone}, given option (D) is Eplerenone. We refer to this prompt as \OpenEliminative\ prompt.

\subsection{Clinical Reasoning Prompt}

In real clinical scenarios, there are no options to choose from. The clinician uses the medical history to form a mental structure for differential diagnosis, performs examinations, adds or deletes the potential diagnosis based on contextual information, and then ultimately takes into consideration laboratory investigations to finally reach the diagnosis. 

We propose a clinical reasoning prompt (\OpenClinicR) that mimics this strategy to reach a final answer. We follow a few-shot approach using CoT, where the CoT would first copy a pertinent piece of patient history from the question, add potential diagnosis based on it, continue the process of adding one piece of patient history at a time, add or delete potential diagnosis, reach the final diagnosis and end with \textit{The answer is $\$answer$}. We also create an MCQ version of the prompt (\MCQClinicR), by replacing the answer with the correct option. For example, \textit{The answer is (D)}.

\subsection{Forward-Backward Approach}

We observed that LLMs mostly generate correct answers when prompted with the \OpenEliminative\ prompt. However, in instances where the model fails to produce the right response, we've noticed that when sampling four completions, one of them is often right. To capitalize on this strength, we propose a forward-backward approach.

This approach has two steps. In the forward step, we use the \OpenClinicR\ prompt and sample multiple possible answers (or) options. We then use simple lexical similarity metrics to remove duplicate options and retain the top-k options with the highest probabilities. In the backward step, we use an MCQ prompt to select one of the options as the answer.

\subsubsection{Verifier}
For the backward step, we propose to use a trained verifier, that predicts the likelihood of the option being an answer given the question. During inference, each option is passed through the verifier and the most likely option is chosen as the answer.

\vspace{0.2ex}
\noindent\textbf{Training:}
We create the training data for the verifier using the \MedQAOriginal\ dataset. For each question $q_i$ in the dataset, we create four examples, each corresponding to a candidate answer $o_i^j$, where $1 \leq j \leq 4$. The correct candidate is set to output 1 (i.e., the likelihood of the candidate being correct is 1) and all the incorrect candidates are set to output 0.

\begin{figure}[t]
   \centering
   \includegraphics[width=0.5\textwidth]{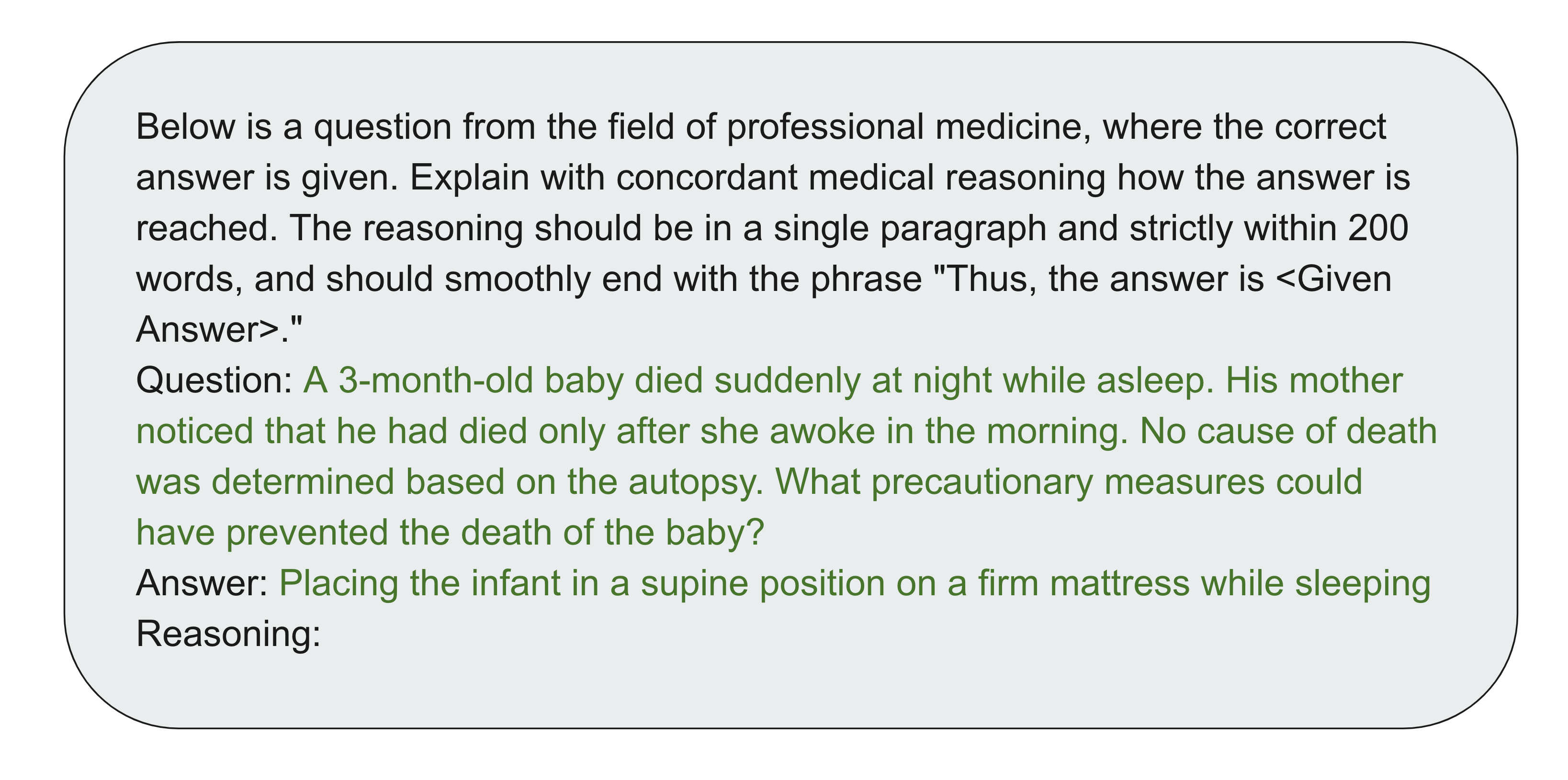}
   \caption{Prompt to generate CoT reasoning for the Verifier training data.}
   \label{fig:verifier_reasoning_generation}
\end{figure}

In the forward step, in addition to the final answer, a CoT reasoning was also generated by the \OpenEliminative\ prompt. In order to better utilize this CoT reasoning, we also generated CoT reasoning $r_i^j$ for all the $(q_i, o_i^j)$ pair and created augmented train examples of the form $(q_i, r_i^j, o_i^j)$. $r_i^j$ were created using the prompt shown in Figure \ref{fig:verifier_reasoning_generation}. We randomly sampled 50 examples for each label and assessed the correctness of the generated CoT reasonings. The results showed that these samples were around 97\% accurate.

In Figure \ref{fig:promptvsDatasets}, we summarize the two prompting strategies, namely \OpenEliminative{} and \OpenClinicR{} used for generating the answers in the open setting. We also show how the two strategies can also be used for the MCQ setting as well. Both prompt strategies use 5-shot examples in the prompt.

\subsection{5-shot examples}
The 5-shot examples were crafted by medical professionals, following a structured clinical reasoning process. First, symptoms in the question were analyzed to identify potential diagnoses, with reasoning applied to narrow down options. As additional signs and test results were introduced, the diagnoses were updated, and reasoning was refined incrementally. This approach mirrored the real-world diagnostic process, incorporating logic at each step. The prompts highlighted this reasoning, ensuring the model replicated not just the final answer but also the diagnostic pathway, enhancing the accuracy and relevance of responses. \\

\noindent \textbf{Implementation Details}\\
We trained two variants of the verifier. One was trained using the Llama-2-7B-chat model as base and the other was trained using the Llama-2-70B-chat model as the base. we added a linear head on each model to train the reward {model}\footnote{We used a learning rate of $5e^{-5}$, with a batch size of 2 and gradient accumulation steps set to 16}.
We apply {Low-Ranked Adaptation (LoRA){\footnote{Lora Parameters: r=16, lora\_alpha =16}} for fine-tuning purpose. The model was trained on a reward modeling loss from \cite{wang2024secrets}.
Optimizer used was $Adam W$. We used 1 Nvidia A100 80GB GPU for training, which took around 72 hrs. For 70B-chat model, we had to use 4 Nvidia A100 80GB GPUs, and train it for 72 hrs.

\section{Experiments}

\subsection{\MCQEliminative{}  {\em vs} \MCQClinicR{} on \MedQAOriginal{} dataset}

We first examine the effectiveness of the clinical reasoning prompting strategy in the MCQ setup. Its underlying prospective incremental reasoning structure typically follows the line of reasoning employed by medical professionals (Appendix A). Along with this, we compare responses selected using the \MCQEliminative{} prompt. The reasoning used in this prompt is more eliminative in nature. It is biased by the options available, rather than using the clinical flow of argument. In Figure~\ref{fig: codex_vs_MedCodex_responses}, we see that the \MCQEliminative{} prompt recites the entire clinical context in one go.

Table \ref{tab:MedQA-original-all} shows the accuracy achieved by the two prompting strategies using the 1273 questions from the test set on the Llama-2-7B-Chat and Llama-2-70B-Chat models. We see that the performance of both the models are quite close to each other, with the \MCQClinicR{} slightly lower than \MCQEliminative{}. This indicates that the clinical reasoning does not have any advantage when the options are already available as a part of the question.

\begin{table}[h!]
\centering
{\small
\begin{tabular}{p{0.65\linewidth} | p{0.25\linewidth}}
\toprule
\textbf{Methods} & \textbf{Accuracy} \\
\midrule
Llama-2-7B ($\MCQEliminative$) & 41\% \\
Llama-2-7B ($\MCQClinicR$) & 40.6\% \\
\midrule
Llama-2-70B ($\MCQEliminative$) & 52\% \\
Llama-2-70B ($\MCQClinicR$) & 50\% \\
\bottomrule
\end{tabular}}
\caption{\label{tab:MedQA-original-all} $\MCQClinicR$ vs. $\MCQEliminative$ on Llama-2 7B-Chat and 70B-Chat models}
\end{table} 

\subsection {\OpenEliminative{} {\em vs.} \OpenClinicR{} on \MedQAOpen{} and \ClinicianCases} 

This experiment was performed to measure the ability of both the prompting strategies to handle open-ended medical questions which are similar to in-clinic scenarios. This is done by comparing the results of both prompting strategies by generating outputs with Llama-2-7B-chat and 70B-chat models.

We randomly select 500 questions from \MedQAOpen{} and all the 25 questions from \ClinicianCases{} dataset for this experiment. As the evaluation is not as straightforward as the MCQ setup, we hired eight final year Medical School students to evaluate the answers generated for the open-ended medical questions. The reasoning quality and final answer were assessed on a 3-point Likert scale  \citep{batterton2017likert}, namely Agree, Neutral, and Disagree.

\noindent \textbf{Observations}: The \OpenClinicR{} prompting strategy received an agreeable assessment on 83\% and 87\% on \MedQAOpen{} dataset on Llama-2-7B-chat and 70B-chat models respectively. Whereas, \OpenEliminative{} prompting strategy received an agreeable assessment of 56\% and 84\% on Llama-2-7B-chat and 70B-chat models respectively. Please refer to Figure \ref{fig:Llama-2-7b-chat-3pt} and Figure \ref{fig:Llama-2-70b-chat-3pt} for the complete breakdown. This clearly shows that \OpenClinicR{} is far superior to \OpenEliminative{} when prompting the 7B model and slightly superior when prompting the 70B model.

On the \ClinicianCases{} dataset, \OpenClinicR{} achieves 88\% and 82\% agreeable assessment with Llama-2-7B-chat and 70B-chat with, whereas \OpenEliminative{} achieves 90\% and 86\% respectively. Please refer to Figure \ref{fig:Llama-2-7b-chat-3pt-real} and Figure \ref{fig:Llama-2-70b-chat-3pt-real}) for more details.

\subsection{Forward-Backward using \OpenClinicR{} and \MCQEliminative{}}
\label{subsection:Forward-backward}
Questions in the \MedQAOpen{} dataset are open-ended and do not have multiple options - similar to a real-world problem setting. In this setup, we utilize the \OpenClinicR{} prompting strategy to sample several options for each question, thereby mimicking the \MedQAOriginal{} dataset format. The question followed by generated options is prompted to the LLM using the \MCQEliminative{} prompt. This will result in the LLM selecting the most appropriate option. To create a set of 4 options, we sampled 10 unique options. Filtered options that have high lexical overlap (to avoid selecting repetitive options). We then selected the top 4 options based on the likelihood assigned by the LLM.

Similar to the previous setup, we evaluate via medical experts the reasoning quality and the final answer on a 3-point Likert scale on both the \MedQAOpen{} dataset (500 QA pairs) and the \ClinicianCases{} dataset (25 QA pairs).

\noindent \textbf{Observations}: On the \MedQAOpen{} dataset, the medical experts agreed with 81\% and 87\% of the answers predicted by Llama-2-7B-chat and 70B-chat respectively (\OpenClinicR-\MCQEliminative{} approach - {\em c.f.}, Figures~\ref{fig:Llama-2-7b-chat-3pt} and \ref{fig:Llama-2-70b-chat-3pt}), while on the \ClinicianCases{} dataset, the corresponding numbers were 88\% and 84\% (Figures \ref{fig:Llama-2-7b-chat-3pt-real} and \ref{fig:Llama-2-70b-chat-3pt-real}).

\begin{figure}[h!]
   \centering
   \includegraphics[width=\linewidth]{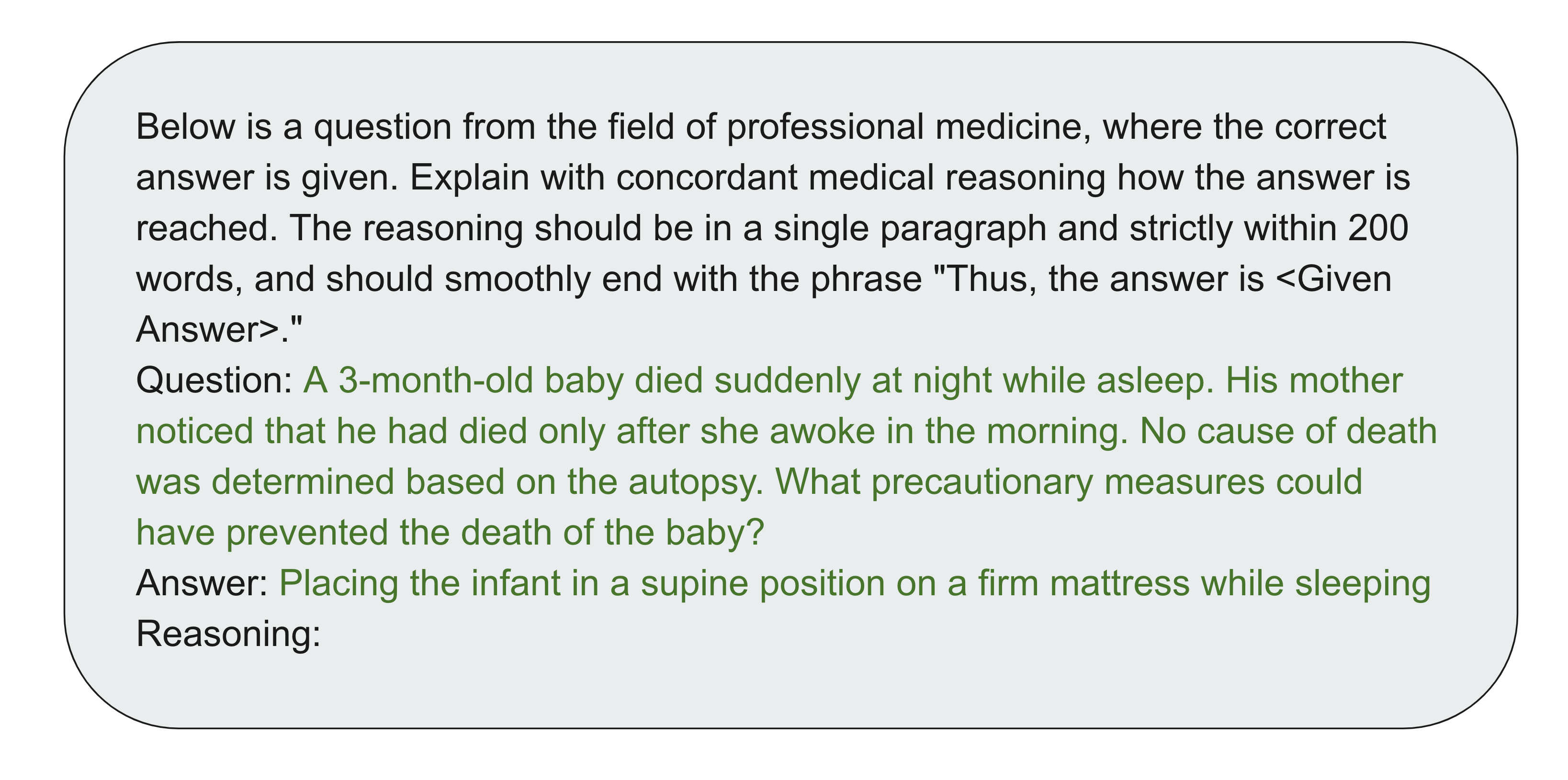}
   \caption{Prompt to generate the reasoning for the Verifier dataset}
   \label{fig:verifier_reasoning_generation}
\end{figure}

\noindent \textbf{Forward-Backward using \OpenClinicR{} and Verifier} \\

We sample forward options as in the previous experiment (\ref{subsection:Forward-backward}), then pass the generated options with their reasoning, as shown in Figure \ref{fig:verifier_input}, to the verifier. We choose the option that gets assigned the highest scalar reward. As before, we select 500 questions from \MedQAOpen{} and evaluate via medical experts the reasoning quality and the final answer on a 3-point Likert scale. 

\noindent \textbf{Observations:} Both Llama-2-7B-chat and 70B-chat achieved scores of 87\% with this method on \MedQAOpen dataset (Figures \ref{fig:Llama-2-7b-chat-3pt} and \ref{fig:Llama-2-70b-chat-3pt}), while the \ClinicianCases{} dataset evaluations reported scores of 90\% and 82\% respectively (Figures~\ref{fig:Llama-2-7b-chat-3pt-real} and \ref{fig:Llama-2-70b-chat-3pt-real}).



\begin{figure}[h!]
   \centering
   \includegraphics[width=0.5\textwidth]{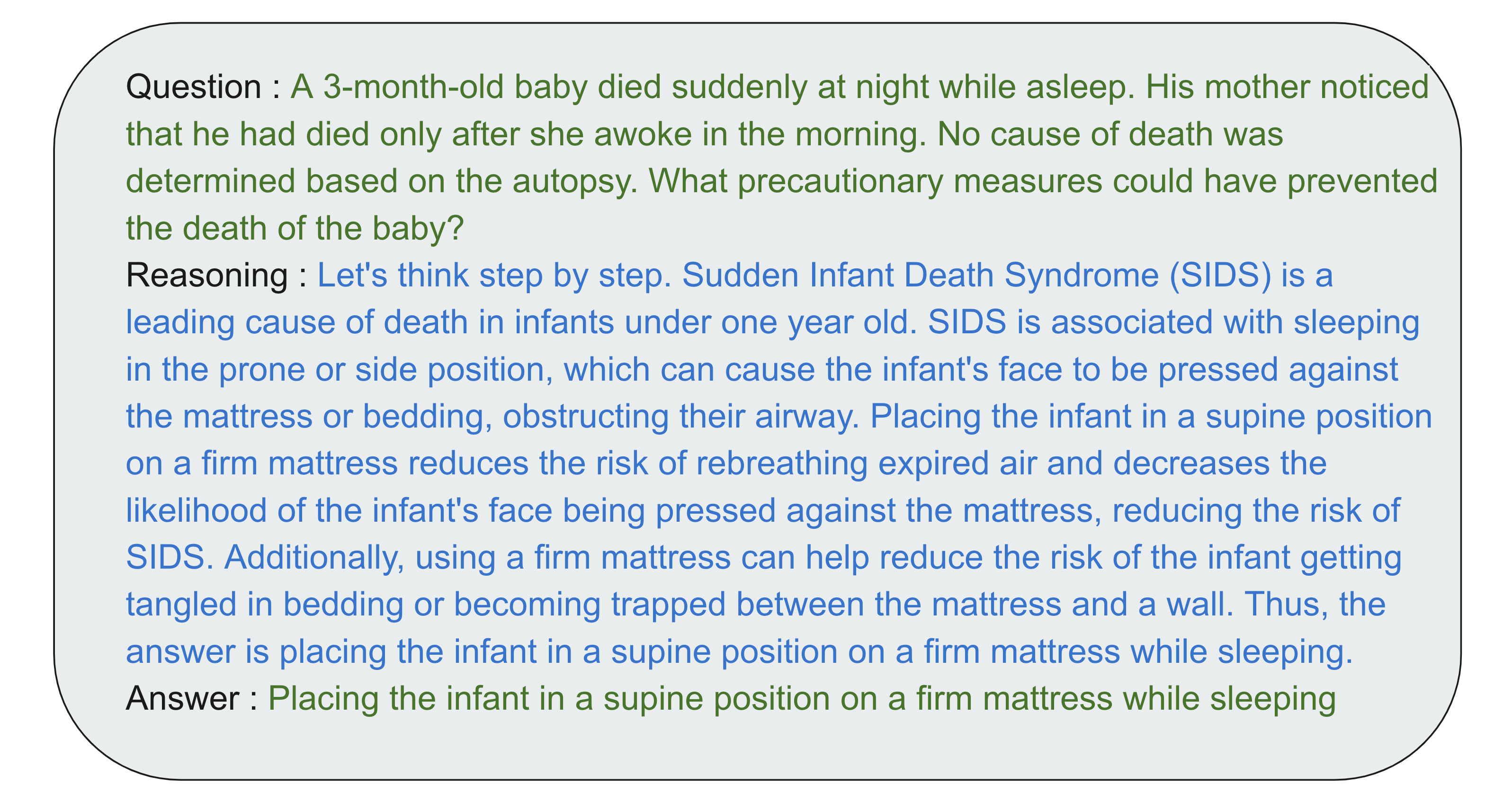}
   \caption{Input prompt to the Verifier}
    \label{fig:verifier_input}
\end{figure}

\begin{figure}[h!]
   \centering
   \includegraphics[width = \linewidth]{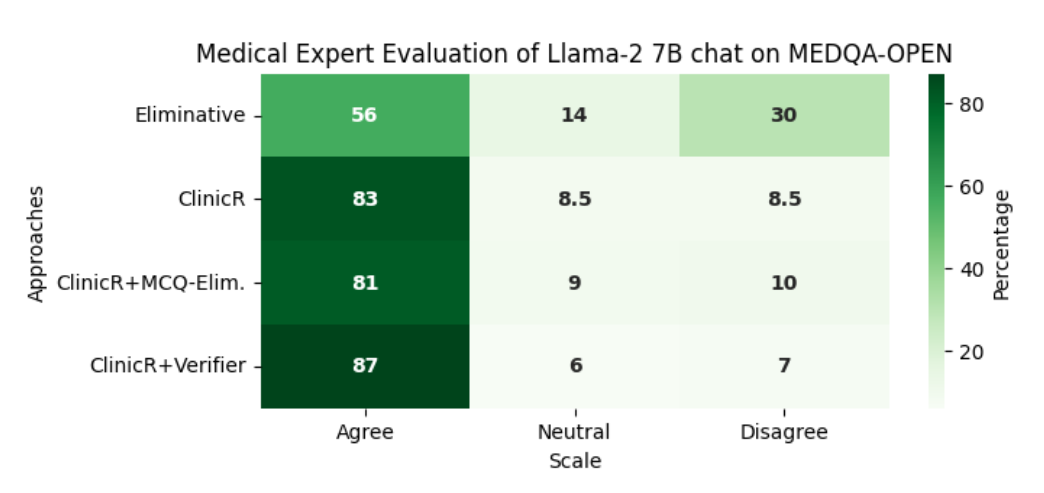}
   \caption{Results of experiments on \MedQAOpen{}  dataset with Llama-2-7B-chat model}
    \label{fig:Llama-2-7b-chat-3pt}
\end{figure}

\vspace{-10pt}

\begin{figure}[h!]
   \centering
   \includegraphics[width = \linewidth]{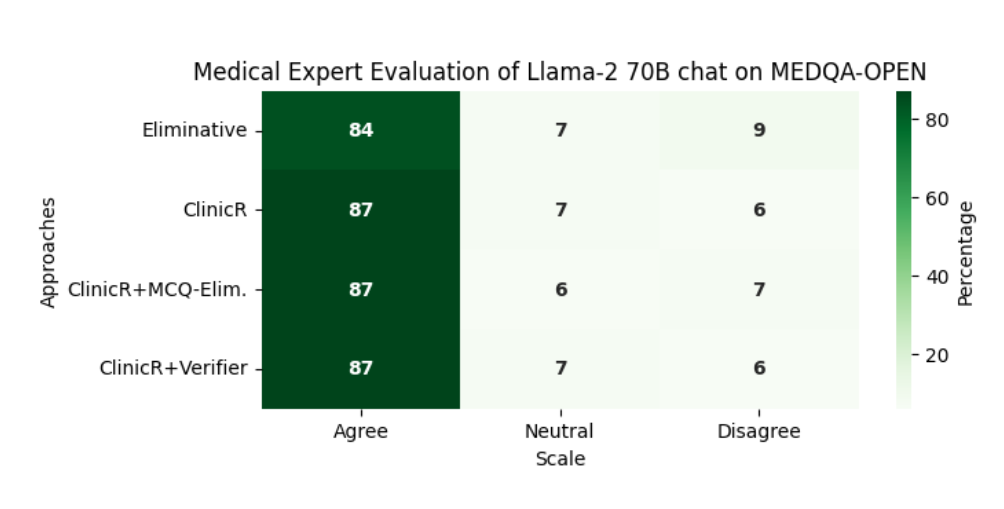}
   \caption{Results of experiments on \MedQAOpen{} dataset with Llama-2-70B-chat model}
    \label{fig:Llama-2-70b-chat-3pt}
\end{figure}

\begin{figure}[h!]
   \centering
   \includegraphics[width = \linewidth]{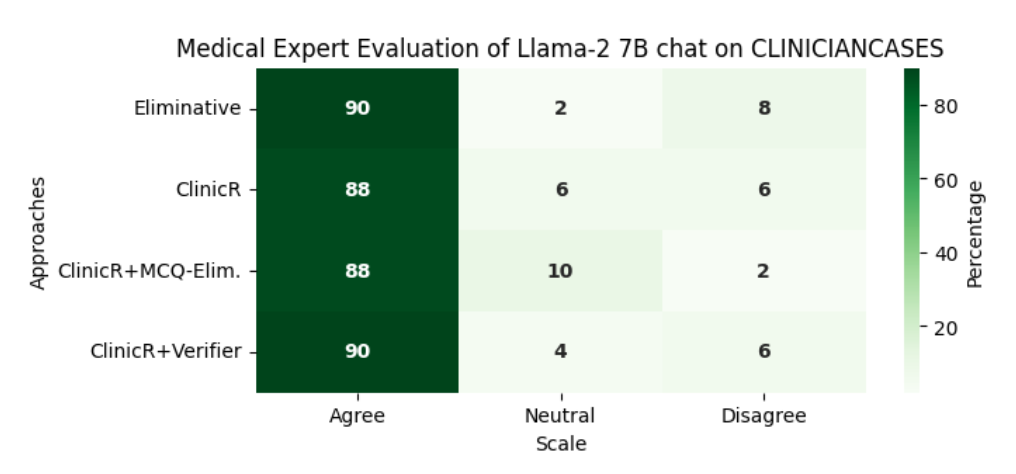}
   \caption{Results of experiments on \ClinicianCases{}  with Llama-2-7B-chat model}
    \label{fig:Llama-2-7b-chat-3pt-real}
\end{figure}
\begin{figure}[h!]
   \centering
   \includegraphics[width = \linewidth]{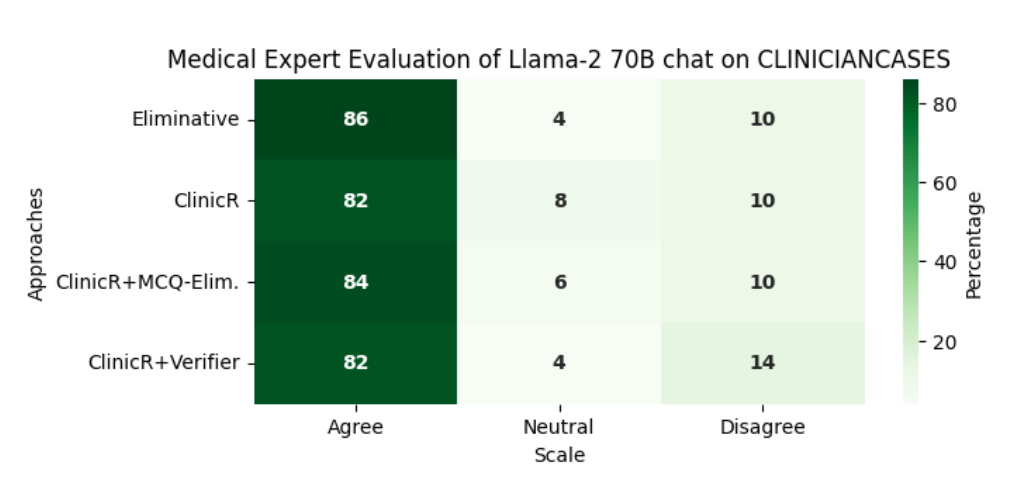}
   \caption{Results of experiments on  \ClinicianCases{}  with Llama-2-70B-chat model}
    \label{fig:Llama-2-70b-chat-3pt-real}
\end{figure}

\vspace{-10pt}

\section*{Interpreting the Experiment Results}


From medical expert evaluations of Llama-2-7B-chat on \MedQAOpen{} (Figure \ref{fig:Llama-2-7b-chat-3pt}), we see that all methods except \OpenEliminative{} perform very well, with \OpenClinicR-Verifier setting giving the highest performance 87\% and the lowest disagreement 7\% in a lower resource setting such as with a Llama-2-7B-chat model. \OpenEliminative{} shows comparatively poorer results 56\% with a high level of disagreements as well 30\%, while \OpenClinicR{} performs on par with the other methods with an agreement of 83\%. We attribute this to the \OpenEliminative's  inability to extract information from the model parameters as effectively as the others due to it's eliminative nature and an absence of options in the open-ended-QA setting. \OpenClinicR{}, on the other hand, is based on an incremental reasoning structure that is designed to draw out inferences from the context and reaches a differential diagnosis.
We see this occurrence disappearing in the evaluations of Llama-2-70B-chat for the \MedQAOpen{} dataset. Here all the methods including \OpenEliminative{}, score really high on the experts' metrics, and can extract the information and reasoning quite effectively from the large knowledge pool of the 70B-chat model. 

With the evaluations of the \ClinicianCases{} dataset on Llama-2-7B-chat model, we see all the methods performing very well with \OpenEliminative{} \ showing slightly higher disagreements (8\%) compared to others. The high performance of all methods is maintained across the 70B-chat model as well, where slightly higher disagreement is displayed by the \OpenClinicR-Verifier setting. 

We hypothesize that the disparity in performance between the MCQ and open-ended datasets arises because the MCQ dataset is specifically designed with distractor options meant to challenge medical students. These distractors can mislead both human participants and models by constraining reasoning within the boundaries of predefined choices. Consequently, the model's ability to generate coherent and comprehensive reasoning is inhibited, as it is forced to navigate between potentially misleading options rather than exploring the full scope of diagnostic possibilities. In contrast, the open-ended dataset allows the model to assess a wider range of possibilities based on the patient’s history and available data, leading to more clinically relevant outputs. This unrestricted format closely mimics real-world medical diagnostics, where clinicians evaluate a broad array of differential diagnoses without being confined to a limited set of options.

\section{Conclusion and Future work}
The incremental reasoning chain of thought prompting (\OpenClinicR) is a novel prompting methodology that follows the usual clinical approach of deciding on real-life clinical settings. We demonstrate that this strategy gives significantly better results than \OpenEliminative, which is primarily designed for MCQ-type questions. Further, we demonstrate that the verifier developed using reasoning performs much better at selecting agreeable responses from the Llama-2 models. Future research will explore the generalizability of this approach by testing on other open-source LLMs. 

This research focuses on improving the quality of responses generated using in-context learning and leveraging the models' parametric knowledge of medicine. We could further aim to integrate these methods with non-parametric knowledge sources, to boost the range of knowledge available, and improve the confidence within the output. For this, methods like Retrieval Augmented Generation (RAG), and knowledge graph integration could be explored.

\section{Limitations:}
This paper does not train an LLM from scratch and only leverages pre-trained models. The quality of the model response depends on the quality of the forward prompt. The current process has been only demonstrated on Llama-2 models and needs to be tested on other models to demonstrate generalizability across models.

\section{Ethics Statement}

The proposed research has been designed with a strong emphasis on ethical considerations.  This study utilizes a de-identified medical question and answer dataset (\MedQAOpen). No human subjects were involved. All the methods were evaluated for their performance on open-source datasets, but did not interact with actual patients for diagnosis. The use of \OpenEliminative{}, a CoT-driven prompt, is designed to mirror the incremental reasoning process in medical diagnosis, ensuring that the decision-making process is transparent and explainable. The utilization of a reward model mechanism for response verification ensures that the system is held accountable for its decisions, further emphasizing the importance of ethical considerations in AI Research, especially in medicine. Future work involving real-world clinical applications would require a separate ethics review and informed consent procedures.

\section*{Acknowledgements}

\bibliography{main}
\newpage
\onecolumn
\appendix


\section{Appendix 1: Prompts used in the research}
\allowdisplaybreaks
\noindent

\noindent This section provides the 5-shot prompts for \MCQClinicR{}, \OpenClinicR{},  \MCQEliminative{}, and \OpenEliminative{} strategies.

\subsection{\MCQClinicR{} Prompt}

\noindent The following are multiple choice questions (with answers) about professional medicine. \\

\noindent Q: A 4670-g (10-lb 5-oz) male newborn is delivered at term to a 26-year-old woman after prolonged labor. Apgar scores are 9 and 9 at 1 and 5 minutes. Examination in the delivery room shows swelling, tenderness, and crepitus over the left clavicle. There is decreased movement of the left upper extremity. Movement of the hands and wrists are normal. A grasping reflex is normal in both hands. An asymmetric Moro reflex is present. The remainder of the examination shows no abnormalities and an anteroposterior x-ray confirms the diagnosis. Which of the following is the most appropriate next step in management?
(A) Physical therapy (B) Pin sleeve to the shirt (C) Splinting of the arm (D) MRI of the clavicle \\

\noindent A: Let's think step-by-step. Prolonged labour would mean it would be a difficult child birth. Apgar score is 9 and does not drop after repeated examination which means that there is no respiratory or neurological problem. Clavicular crepitus, tenderness and swelling indicates that there could be clavicular injury since this is common in difficult child birth. Decreased movement in the upper extremity supports some injury indication in the upper limb. Hand and wrist movement being normal as well as grasping movement being normal means that there is no neurological deficit and the problem lies with the arm or shoulder and not the forearm or wrist. Grasping reflex in both hands again indicates that there is no deficit between communication from the spinal cord. An assymetric Moros reflex indicates some local injury which is due to either damage to a peripheral nerve, cervical cord or fracture of the clavicle. However, presence of clavicular crepitus and tenderness indicates towards clavicular fracture and not other conditions. So, the most probable diagnosis is clavicular fracture. Since Clavicular fractures heal very quickly in newborns, hence only restraining is necessary of the injured arm. Thus, the first line of treatment is pinning sleeve to the shirt. The answer is (B). \\

\noindent Q: A previously healthy 32-year-old man is brought to the emergency department by his girlfriend after having a seizure. Earlier that day, he also experienced a nosebleed that took 30 minutes to stop when applying pressure. He has had no sick contacts or history of epilepsy or other seizure disorder. He does not take any medications. His temperature is 39.1 °C (102.4 °F), pulse is 106/min, respirations are 26/min, and blood pressure is 128/70 mm Hg. He is confused and disoriented. Examination shows pallor and scattered petechiae over the trunk and arms. The neck is supple, and neurological examination is otherwise within normal limits. Laboratory studies show:
Hemoglobin 9 g/dL
Leukocyte count 8,200/mm3
Platelet count 34,000/mm3
Prothrombin time 13 seconds
Partial thromboplastin time 30 seconds
Fibrin split products negative
Serum
Creatinine 2.9 mg/dL
Bilirubin
Total 3.2 mg/dL
Direct 0.4 mg/dL
Lactate dehydrogenase 559 U/L
A peripheral blood smear shows numerous schistocytes. Which of the following is the most appropriate next step in management?"
(A) Transfusion of packed red blood cells (B) Plasma exchange therapy (C) Platelet transfusion (D) Intravenous tranexamic acid \\

\noindent A: Let's think step-by-step. Patient arrives with seizures but has no previous history of seizures, so previous neurological problems are less likely. Presence of nose bleed that did not stop for 30 mins indicates that there can be a clotting problem which could be either due to platelet insufficiency or clotting factor problems. There is fever along with seizures which could indicate meningitis. However, neck being supple, meaning not rigid goes against meningitis. Patient is not on any medication which rules out involuntary overdose. Pulse is slightly elevated which indicates that there is hemodynamic instability. Patient is confused and disoriented which could be post seizure effect. There is history of nose bleed as well as petechiae all over the body indicates a problem of clotting which is supported by low platelet from the lab tests. There is fever but leucocyte count is normal which indicates against viral or bacterial infection. The creatinine level is elevated which indicates that the kidneys are being affected. Bilirubin level is elevated, but direct bilirubin is normal indicating that there is excessive bilirubin production potentially due to increased haemolysis which is supported by low Haemoglobin level. Presence of haemolysis is supported by schistocytes observed in peripheral blood smear. However, fragmented blood cells are seen and platelet count is low. All of this points towards a diagnosis of thrombocytopenic purpura. Hence, plasma exchange therapy is the correct answer. The answer is (B). \\

\noindent Q: A 50-year-old man comes to the physician because of a 6-month history of difficulties having sexual intercourse due to erectile dysfunction. He has type 2 diabetes mellitus that is well controlled with metformin. He does not smoke. He drinks 5–6 beers daily. His vital signs are within normal limits. Physical examination shows bilateral pedal edema, decreased testicular volume, and increased breast tissue. The spleen is palpable 2 cm below the left costal margin. Abdominal ultrasound shows an atrophic, hyperechoic, nodular liver. An upper endoscopy is performed and shows dilated submucosal veins 2 mm in diameter with red spots on their surface in the distal esophagus. Therapy with a sildenafil is initiated for his erectile dysfunction. Which of the following is the most appropriate next step in management of this patient's esophageal findings?
(A) Injection sclerotherapy (B) Nadolol therapy (C) Isosorbide mononitrate therapy (D) Transjugular intrahepatic portosystemic shunt \\

\noindent A: Let's think step-by-step. The patient has a history of erectile dysfunction during sexual intercourse. The patient also has a history of Diabetes but it is well controlled with metformin therapy. However, the patient has a long history of alcohol abuse as indicated by 5 beers daily. There is decreased testicular volume and increased breast tissue which could indicates androgenic insufficiency. There is also pedal oedema which can be due to uncontrolled blood pressure or due to liver dysfunction. The spleen is enlarged as indicated by being palpable 2cm below left costal margin which indicates increased portal pressure. Abdominal ultrasonography indicates pressure of alcoholic liver cirrhosis as indicated by atrophic nodular liver. The enlarged spleen can then be potentially due to portal hypertension secondary to liver cirrhosis. Liver cirrhosis also causes decreased protein synthesis which causes pedal oedema due to decreased blood osmotic pressure. Liver cirrhosis also causes increased estrogen to testosterone ratio which results in the physical manifestation of decreased testicular volume and increased breast tissue as well as erectile dysfunction. Endoscopy shows dilated submucosal veins 2 mm in diameter with red spots on their surface in the distal esophagus which again indicates portal hypertension. The diagnosis of this condition is portal hypertension secondary to liver cirrhosis due to chronic alcoholism. Primary therapy to treat portal hypertension here is to decrease portal pressure which is done by administering beta blockers. So, Nadolol therapy should be initiated. The answer is (B).  \\

\noindent Q: A 24-year-old primigravid woman at 38 weeks' gestation comes to the physician for a prenatal visit. At the last two prenatal visits, transabdominal ultrasound showed the fetus in breech presentation. She has no medical conditions and only takes prenatal vitamins. Her pulse is 95/min, respirations are 16/min, and blood pressure is 130/76 mm Hg. The abdomen is soft and nontender; no contractions are felt. Pelvic examination shows a closed cervical os and a uterus consistent with 38 weeks' gestation. The fetal rate tracing shows a baseline heart rate of 152/min and 2 accelerations over 10 minutes. Repeat ultrasound today shows a persistent breech presentation. The patient states that she would like to attempt a vaginal delivery. Which of the following is the most appropriate next step in management?
(A) Recommend cesarean section (B) Observe until spontaneous labor (C) Repeat ultrasound in one week (D) Offer external cephalic version \\

\noindent A: Let's think step-by-step. This is the first pregnancy of the woman and she is close to term as indicated by 38 weeks of gestation. Previous two abdominal USG’s indicated a breech presentation of the foetus. The vitals are normal, i.e pulse and respiratory rate is normal and blood pressure is normal indicating not preeclampsia so no immediate threat to the mother’s health and there are no contractions indicating that while the mother is close to term she is not in labour. Also, the abdomen is soft and tender and there are no contractions which indicates that the woman is not in labour. Also, the pelvic examination shows that the cervical os is closed again indicating that the mother is not in labour. The foetal heart rate is 152/min and 2 accelerations over 10 mins again indicating that mother is not in labour and the foetus is healthy. An ultrasound done today confirms the breech presentation of the foetus. However, since the woman is close to term it is unlikely that the featus will turn by itself and will continue to stay in breech position. While a C-section is an alternative in this situation, since the mother has indicated that she would want to attempt a normal vaginal delivery, an external cephalic version can be offered to the woman. The answer is (D). \\

\noindent Q: A 4-year old boy is brought to the emergency department with fever, painful swallowing, headache, and neck spasm that began shortly after waking up. He has had a sore throat over the last week that acutely worsened this morning. He has no history of serious illness and takes no medications. He lives at home with his mother. His older brother has asthma. His immunizations are up-to-date. He appears acutely ill. His temperature is 38.4°C (101.2°F), pulse is 95/min, respirations are 33/min, and blood pressure is 93/60 mm Hg. Examination shows drooling. The neck is stiff and extension is limited. Respirations appear labored with accessory muscle use. Inspiratory stridor is heard on auscultation of the chest. Cardiac examination shows no abnormalities. Oropharyngeal examination shows a bulge in the posterior pharyngeal wall. Intravenous access is obtained and laboratory studies are ordered. Which of the following is the most appropriate next step in the management of this patient?
(A) Endotracheal intubation (B) IV antibiotics (C) Blood cultures (D) IV corticosteroids \\

\noindent A: Let's think step-by-step. The 4-year-old boy has fever 38.4 C, painful swallowing and headache which indicates upper respiratory tract infection. There is neck spasm and fever which could indicate meningitis. Sore throat presence again indicates upper respiratory tract infection which is worsening since this morning. No history of medication indicates that there is less likelihood of accidental overdose. Brother has a history of asthma which increases his own risk of autoimmune conditions. Neck is stiff, there is drooling which could indicates neurological affliction which points towards meningitis. This could be meningitis since meningococcal vaccine is given around 10-12 years. Drooling could also indicate extreme difficulty in swallowing. Respiration is laboured with use of accessory muscle indicates extreme respiratory distress. No cardiac abnormalities so, respiratory distress is primarily respiratory in origin. There is a huge budge in the posterior pharyngeal wall which could be of acute origin since the sore throat is acutely worsened since morning. This could be either meningitis secondary to upper respiratory tract infection but the first step is to secure the airway since there is severe respiratory distress. So endotracheal intubation should be done immediately. The answer is (A). \\

\subsection{\OpenClinicR{} Prompt}

\noindent Use just the given patient history to answer the question. Do not assume any further information about the patient. Strictly Limit your response to 200 words. \\

\noindent Q: A 4670-g (10-lb 5-oz) male newborn is delivered at term to a 26-year-old woman after prolonged labor. Apgar scores are 9 and 9 at 1 and 5 minutes. Examination in the delivery room shows swelling, tenderness, and crepitus over the left clavicle. There is decreased movement of the left upper extremity. Movement of the hands and wrists are normal. A grasping reflex is normal in both hands. An asymmetric Moro reflex is present. The remainder of the examination shows no abnormalities and an anteroposterior x-ray confirms the diagnosis. What is the most suitable course of action to manage this case? \\

\noindent A: Let's think step-by-step. Prolonged labour would mean it would be a difficult child birth. Apgar score is 9 and does not drop after repeated examination which means that there is no respiratory or neurological problem. Clavicular crepitus, tenderness and swelling indicates that there could be clavicular injury since this is common in difficult child birth. Decreased movement in the upper extremity supports some injury indication in the upper limb. Hand and wrist movement being normal as well as grasping movement being normal means that there is no neurological deficit and the problem lies with the arm or shoulder and not the forearm or wrist. Grasping reflex in both hands again indicates that there is no deficit between communication from the spinal cord. An assymetric Moros reflex indicates some local injury which is due to either damage to a peripheral nerve, cervical cord or fracture of the clavicle. However, presence of clavicular crepitus and tenderness indicates towards clavicular fracture and not other conditions. So, the most probable diagnosis is clavicular fracture. Since Clavicular fractures heal very quickly in newborns, hence only restraining is necessary of the injured arm. Thus, the first line of treatment is pinning sleeve to the shirt.
Answer: pinning sleeve to the shirt. \\

\noindent Use just the given patient history to answer the question. Do not assume any further information about the patient. Strictly Limit your response to 200 words. \\

\noindent Q: A previously healthy 32-year-old man is brought to the emergency department by his girlfriend after having a seizure. Earlier that day, he also experienced a nosebleed that took 30 minutes to stop when applying pressure. He has had no sick contacts or history of epilepsy or other seizure disorder. He does not take any medications. His temperature is 39.1 °C (102.4 °F), pulse is 106/min, respirations are 26/min, and blood pressure is 128/70 mm Hg. He is confused and disoriented. Examination shows pallor and scattered petechiae over the trunk and arms. The neck is supple, and neurological examination is otherwise within normal limits. Laboratory studies show:
Hemoglobin 9 g/dL
Leukocyte count 8,200/mm3
Platelet count 34,000/mm3
Prothrombin time 13 seconds
Partial thromboplastin time 30 seconds
Fibrin split products negative
Serum
Creatinine 2.9 mg/dL
Bilirubin
Total 3.2 mg/dL
Direct 0.4 mg/dL
Lactate dehydrogenase 559 U/L
A peripheral blood smear shows numerous schistocytes. What is the most appropria    te next step in the management of this patient? \\

\noindent A: Let's think step-by-step. Patient arrives with seizures but has no previous history of seizures, so previous neurological problems are less likely. Presence of nose bleed that did not stop for 30 mins indicates that there can be a clotting problem which could be either due to platelet insufficiency or clotting factor problems. There is fever along with seizures which could indicate meningitis. However, neck being supple, meaning not rigid goes against meningitis. Patient is not on any medication which rules out involuntary overdose. Pulse is slightly elevated which indicates that there is hemodynamic instability. Patient is confused and disoriented which could be post seizure effect. There is history of nose bleed as well as petechiae all over the body indicates a problem of clotting which is supported by low platelet from the lab tests. There is fever but leucocyte count is normal which indicates against viral or bacterial infection. The creatinine level is elevated which indicates that the kidneys are being affected. Bilirubin level is elevated, but direct bilirubin is normal indicating that there is excessive bilirubin production potentially due to increased haemolysis which is supported by low Haemoglobin level. Presence of haemolysis is supported by schistocytes observed in peripheral blood smear. However, fragmented blood cells are seen and platelet count is low. All of this points towards a diagnosis of thrombocytopenic purpura. Hence, plasma exchange therapy is the correct answer.
Answer: plasma exchange therapy. \\

\noindent Use just the given patient history to answer the question. Do not assume any further information about the patient. Strictly Limit your response to 200 words. \\

\noindent Q: A 67-year-old man comes to the emergency department for the evaluation of two episodes of red urine since this morning. He has no pain with urination. He reports lower back pain and fever. Six months ago, he was diagnosed with osteoarthritis of the right knee that he manages with 1–2 tablets of ibuprofen per day. He has smoked one pack of cigarettes daily for the past 45 years. He does not drink alcohol. His temperature is 38.5°C (101.3°F), pulse is 95/min, and blood pressure is 130/80 mm Hg. Physical examination shows faint, diffuse maculopapular rash, and bilateral flank pain. The remainder of the examination shows no abnormalities. What is the most likely diagnosis for this patient? \\

\noindent A: Let's think step-by-step. The patient has a history of erectile dysfunction during sexual intercourse. The patient also has a history of Diabetes but it is well controlled with metformin therapy. However, the patient has a long history of alcohol abuse as indicated by 5 beers daily. There is decreased testicular volume and increased breast tissue which could indicates androgenic insufficiency. There is also pedal oedema which can be due to uncontrolled blood pressure or due to liver dysfunction. The spleen is enlarged as indicated by being palpable 2cm below left costal margin which indicates increased portal pressure. Abdominal ultrasonography indicates pressure of alcoholic liver cirrhosis as indicated by atrophic nodular liver. The enlarged spleen can then be potentially due to portal hypertension secondary to liver cirrhosis. Liver cirrhosis also causes decreased protein synthesis which causes pedal oedema due to decreased blood osmotic pressure. Liver cirrhosis also causes increased estrogen to testosterone ratio which results in the physical manifestation of decreased testicular volume and increased breast tissue as well as erectile dysfunction. Endoscopy shows dilated submucosal veins 2 mm in diameter with red spots on their surface in the distal esophagus which again indicates portal hypertension. The diagnosis of this condition is portal hypertension secondary to liver cirrhosis due to chronic alcoholism. Primary therapy to treat portal hypertension here is to decrease portal pressure which is done by administering beta blockers. So, Nadolol therapy should be initiated.
Answer: Nadolol Therapy. \\

\noindent Use just the given patient history to answer the question. Do not assume any further information about the patient. Strictly Limit your response to 200 words. \\

\noindent Q: A 24-year-old primigravid woman at 38 weeks' gestation comes to the physician for a prenatal visit. At the last two prenatal visits, transabdominal ultrasound showed the fetus in breech presentation. She has no medical conditions and only takes prenatal vitamins. Her pulse is 95/min, respirations are 16/min, and blood pressure is 130/76 mm Hg. The abdomen is soft and nontender; no contractions are felt. Pelvic examination shows a closed cervical os and a uterus consistent with 38 weeks' gestation. The fetal rate tracing shows a baseline heart rate of 152/min and 2 accelerations over 10 minutes. Repeat ultrasound today shows a persistent breech presentation. The patient states that she would like to attempt a vaginal delivery. Describe the next best step for managing this patient. \\

\noindent A: Let's think step-by-step. This is the first pregnancy of the woman and she is close to term as indicated by 38 weeks of gestation. Previous two abdominal USG’s indicated a breech presentation of the foetus. The vitals are normal, i.e pulse and respiratory rate is normal and blood pressure is normal indicating not preeclampsia so no immediate threat to the mother’s health and there are no contractions indicating that while the mother is close to term she is not in labour. Also, the abdomen is soft and tender and there are no contractions which indicates that the woman is not in labour. Also, the pelvic examination shows that the cervical os is closed again indicating that the mother is not in labour. The foetal heart rate is 152/min and 2 accelerations over 10 mins again indicating that mother is not in labour and the foetus is healthy. An ultrasound done today confirms the breech presentation of the foetus. However, since the woman is close to term it is unlikely that the featus will turn by itself and will continue to stay in breech position. While a C-section is an alternative in this situation, since the mother has indicated that she would want to attempt a normal vaginal delivery, an external cephalic version can be offered to the woman.
Answer: offering external cephalic version. \\

\noindent Use just the given patient history to answer the question. Do not assume any further information about the patient. Strictly Limit your response to 200 words. \\

\noindent Q: A 4-year old boy is brought to the emergency department with fever, painful swallowing, headache, and neck spasm that began shortly after waking up. He has had a sore throat over the last week that acutely worsened this morning. He has no history of serious illness and takes no medications. He lives at home with his mother. His older brother has asthma. His immunizations are up-to-date. He appears acutely ill. His temperature is 38.4°C (101.2°F), pulse is 95/min, respirations are 33/min, and blood pressure is 93/60 mm Hg. Examination shows drooling. The neck is stiff and extension is limited. Respirations appear labored with accessory muscle use. Inspiratory stridor is heard on auscultation of the chest. Cardiac examination shows no abnormalities. Oropharyngeal examination shows a bulge in the posterior pharyngeal wall. Intravenous access is obtained and laboratory studies are ordered. Explain the most suitable course of action for managing this patient's condition. \\

\noindent A: Let's think step-by-step. The 4-year-old boy has fever 38.4 C, painful swallowing and headache which indicates upper respiratory tract infection. There is neck spasm and fever which could indicate meningitis. Sore throat presence again indicates upper respiratory tract infection which is worsening since this morning. No history of medication indicates that there is less likelihood of accidental overdose. Brother has a history of asthma which increases his own risk of autoimmune conditions. Neck is stiff, there is drooling which could indicates neurological affliction which points towards meningitis. This could be meningitis since meningococcal vaccine is given around 10-12 years. Drooling could also indicate extreme difficulty in swallowing. Respiration is laboured with use of accessory muscle indicates extreme respiratory distress. No cardiac abnormalities so, respiratory distress is primarily respiratory in origin. There is a huge budge in the posterior pharyngeal wall which could be of acute origin since the sore throat is acutely worsened since morning. This could be either meningitis secondary to upper respiratory tract infection but the first step is to secure the airway since there is severe respiratory distress. So endotracheal intubation should be done immediately.
Answer: endotracheal intubation. \\

\subsection{\MCQEliminative{} Prompt}
The following are multiple choice questions (with answers) about professional medicine. \\

\noindent Q: Four weeks after starting hydrochlorothiazide, a 49-year-old man with hypertension comes to the physician because of muscle cramps and weakness. His home medications also include amlodipine. His blood pressure today is 176/87 mm Hg. Physical examination shows no abnormalities. The precordial leads of a 12-lead ECG are shown. The addition of which of the following is most likely to have prevented this patient's condition?
(A) Torsemide (B) Nifedipine (C) Eplerenone (D) Hydralazine \\

\noindent A: Let's think step by step. The patient has started hydrochlorothiazide. He now presents with muscle cramps and weakness and an ECG that supports the diagnosis of hypokalemia. The addition of (A) Torsemide is a loop diuretic and would likely aggravate the hypokalemia, (B) Nifedipine is a calcium antagonist and would not alleviate the hypocalcemia. (C) Eplerenone is a potassium-sparing diuretic and would likely decrease the chance of hypokalemia. (C) Eplerenone is a potent vasodilator and would not decrease the risk of hypokalemia. The answer is (C). \\

\noindent Q: A 57-year-old woman comes to the emergency department because of severe pain around her right eye, blurred vision in the same eye, and a headache for the past 4 hours. She is nauseous but has not vomited. She can see colored bright circles when she looks at a light source. She is currently being treated for a urinary tract infection with trimethoprim-sulfamethoxazole. She appears uncomfortable. Vital signs are within normal limits. Examination shows visual acuity of 20/20 in the left eye and counting fingers at 5 feet in the right eye. The right eye shows conjunctival injection and edema of the cornea. The right pupil is dilated and fixed. Intravenous analgesia and antiemetics are administered. Which of the following is the most appropriate next step in management?
(A) Perform ultrasound biomicroscopy (B) Perform gonioscopy (C) Perform fundoscopy (D) Administer topical steroids \\

\noindent A: Let's think step by step. The patient has severe pain, blurred vision, and a headache. She is also nauseous but has not vomited. She can see bright colored circles when she looks at a light source. The right eye shows conjunctival injection and edema of the cornea. The right pupil is dilated and fixed. The tentative (most likely) diagnosis, given the patient's symptoms, is acute primary angle-closure glaucoma. Analgesics have been administered. The next step is to confirm the tentative diagnosis using the gold standard test. (A) ultrasound biomicroscopy could be used but are not widely available, (B) Gonioscopy is the gold-standard test to confirm the diagnosis of acute primary angle-closure glaucoma, (C) fundoscopy is not as reliable as gonioscopy, (D) topical steroids is not relevant, the possible medical therapy would be timolol, apraclonidine or pilocarpine. The most appropriate next step in management is to administer intravenous analgesia and antiemetics. The answer is (B). \\

\noindent Q: A 55-year-old man with a history of hypertension and benign prostate hyperplasia presents for follow-up 4 days into the treatment of a urinary tract infection with trimethoprim-sulfamethoxazole. His symptoms have resolved, and he reports no problems with urination, with the exception of a weak urine stream and hesitancy, which he has had for the past 2 years. At the time of this visit, the patient is afebrile; the blood pressure is 130/88 mm Hg and the heart rate is 80/min. There is no flank tenderness. A urinalysis reveals no leukocytes and is negative for esterase. The urinalysis reveals 2 red blood cells (RBCs)/ high power field (HPF), and there are no casts on urinary sediment analysis. The physician, however, notices the following abnormality:
Prior treatment
BUN 12 mg/dL
Creatinine 1.2 mg/dL
Today’s visit
BUN 13 mg/dL
Creatinine 2.1 mg/dL
(A) Reassure the patient, stop trimethoprim-sulfamethoxazole and repeat the measurement in 1–2 weeks (B) Schedule an intravenous pyelography for urinary obstruction (C) Schedule a cystoscopy for urethral obstruction (D) Admit the patient for further management of acute interstitial nephritis \\

\noindent A: Let's think step by step. The patient has a history of hypertension and benign prostate hyperplasia. These are both risk factors for urinary tract infections. The patient is being treated with trimethoprim-sulfamethoxazole for a urinary tract infection. The symptoms have resolved, but the patient has a weak urine stream and hesitancy. The urinalysis reveals 2 red blood cells (RBCs)/ high power field (HPF), and there are no casts on urinary sediment analysis. The physician, however, notices the following abnormality: BUN 12 mg/dL, Creatinine 1.2 mg/dL (prior treatment) vs BUN 13 mg/dL, Creatinine 2.1 mg/dL (today’s visit). The BUN and creatinine levels have increased from the prior visit. This could be due to acute interstitial nephritis. A known but rare side effect of trimethoprim-sulfamethoxazole. Interstitial nephritis should be suspected if there is an elevated creatinine and a urinalysis that shows white cells, white cell casts, and, in some cases, eosinophiluria. Since none of the urinalysis results is present and the patient has no typical symptoms such as fever, rash, or acute interstitial nephritis is an unlikely diagnosis. The answer is (A). \\

\noindent Q: A 32-year-old woman comes to the physician because of weight gain, generalized weakness, and irregular menstrual cycles for the past 16 months. She began having symptoms of insomnia and depression 10 months ago. More recently, she has been having difficulties rising from a chair. She has a 2-year history of hypertension. Current medications include citalopram and hydrochlorothiazide. She is 168 cm (5 ft 6 in) tall and weighs 100 kg (220 lb). BMI is 36 kg/m2. Her pulse is 92/min, respirations are 18/min, and blood pressure is 134/76 mm Hg. She appears tired and has a full, plethoric face and central obesity. Examination of the skin shows violaceous linear striations on her lower abdomen. Two midnight serum cortisol studies show measurements of 288 \textmu g/L and 253 \textmu g/L (N < 90); a 24-hour urinary cortisol measurement was 395 \textmu g (N < 300). Upon follow-up laboratory examination, the patient's serum ACTH levels were also elevated at 136 pg/mL (N = 7–50). Which of the following is the most appropriate next step in evaluation?
(A) MRI of the head with contrast (B) Measure ACTH levels in inferior petrosal sinuses (C) High-dose dexamethasone suppression test (D) CT scan of the abdomen with contrast \\

\noindent A: Let's think step by step. The patient has Cushing's syndrome, which is characterized by weight gain, generalized weakness, and irregular menstrual cycles. She also has symptoms of insomnia and depression, which could be caused by Cushing's syndrome. Additionally, she has difficulties rising from a chair, which could be caused by muscle weakness. The two elevated serum cortisol measurements and the elevated ACTH levels confirm the diagnosis of Cushing's syndrome. A high dose dexamethasone suppression test will help determine if the cause of Cushing's syndrome is Cushing's disease (CS caused by pituitary hypersecretion of ACTH) or ectopic ACTH syndrome (CS caused by nonpituitary ACTH-secreting tumors). Since the diagnosis of Cushing's syndrome is established, the most appropriate next step is to distinguish between Cushing's disease and ectopic ACTH syndrome using the high-dose dexamethasone suppression test. The answer is (C). \\

\noindent Q: A 53-year-old man comes to the physician because of a 3-month history of a nonpruritic rash. He has been feeling more tired than usual and occasionally experiences pain in his wrists and ankles. He does not smoke or drink alcohol. His temperature is 37.6°C (99.7°F), pulse is 98/min, respirations are 18/min, and blood pressure is 130/75 mm Hg. Physical examination shows multiple, erythematous, purpuric papules on his trunk and extremities that do not blanch when pressed. The remainder of the examination shows no abnormalities. The patient's hemoglobin is 14 g/dL, leukocyte count is 9,500/mm3, and platelet count is 228,000/mm3. Urinalysis and liver function tests are within normal limits. The test for rheumatoid factor is positive. Serum ANA is negative. Serum complement levels are decreased. Serum protein electrophoresis and immunofixation shows increased gammaglobulins with pronounced polyclonal IgM and IgG bands. Testing for cryoglobulins shows no precipitate after 24 hours. Chest x-ray and ECG show no abnormalities. Which of the following is the most appropriate next step in management?
(A) Rapid plasma reagin test (B) Hepatitis C serology (C) pANCA assay (D) Bence Jones protein test \\

\noindent A: Let's think step by step. The patient has a rash, fatigue, and pain in his wrists and ankles. He has normal hemoglobin (normal range 13.8-17.2 g/dL), normal leukocyte count (normal range 4500-11000 wbc/microliter), and normal platelet count (normal range 150-450 platelets/microliter). His serum complement levels are decreased. His serum protein electrophoresis and immunofixation show increased gammaglobulins with pronounced polyclonal IgM and IgG bands. The cryoglobulin precipitation test is normal, ruling out cryoglobulinemia. Serum ANA is negative, so Lupus is less likely. Physical examination and symptoms are consistent with possible non-cryoglobulinemic vasculitis. The most likely associated diagnosis is Hepatitis C. The answer is (B). \\

\subsection{\OpenEliminative{} Prompt}

\noindent Q: Four weeks after starting hydrochlorothiazide, a 49-year-old man with hypertension comes to the physician because of muscle cramps and weakness. His home medications also include amlodipine. His blood pressure today is 176/87 mm Hg. Physical examination shows no abnormalities. The precordial leads of a 12-lead ECG are shown. The addition of what is most likely to have prevented this patient's condition? \\

\noindent A: Let's think step by step. The patient has started hydrochlorothiazide. He now presents with muscle cramps and weakness and an ECG that supports the diagnosis of hypokalemia. The addition of Torsemide is a loop diuretic and would likely aggravate the hypokalemia, Nifedipine is a calcium antagonist and would not alleviate the hypocalcemia. Eplerenone is a potassium-sparing diuretic and would likely decrease the chance of hypokalemia. Eplerenone is a potent vasodilator and would not decrease the risk of hypokalemia.
Answer: Eplerenone. \\

\noindent Q: A 57-year-old woman comes to the emergency department because of severe pain around her right eye, blurred vision in the same eye, and a headache for the past 4 hours. She is nauseous but has not vomited. She can see colored bright circles when she looks at a light source. She is currently being treated for a urinary tract infection with trimethoprim-sulfamethoxazole. She appears uncomfortable. Vital signs are within normal limits. Examination shows visual acuity of 20/20 in the left eye and counting fingers at 5 feet in the right eye. The right eye shows conjunctival injection and edema of the cornea. The right pupil is dilated and fixed. Intravenous analgesia and antiemetics are administered. What is the most appropriate next step in management? \\

\noindent A: Let's think step by step. The patient has severe pain, blurred vision, and a headache. She is also nauseous but has not vomited. She can see bright colored circles when she looks at a light source. The right eye shows conjunctival injection and edema of the cornea. The right pupil is dilated and fixed. The tentative (most likely) diagnosis, given the patient's symptoms, is acute primary angle-closure glaucoma. Analgesics have been administered. The next step is to confirm the tentative diagnosis using the gold standard test. Ultrasound biomicroscopy could be used but are not widely available, Gonioscopy is the gold-standard test to confirm the diagnosis of acute primary angle-closure glaucoma, fundoscopy is not as reliable as gonioscopy, topical steroids is not relevant, the possible medical therapy would be timolol, apraclonidine or pilocarpine. The most appropriate next step in management is to administer intravenous analgesia and antiemetics.
Answer: Gonioscopy. \\

\noindent Q: A 55-year-old man with a history of hypertension and benign prostate hyperplasia presents for follow-up 4 days into the treatment of a urinary tract infection with trimethoprim-sulfamethoxazole. His symptoms have resolved, and he reports no problems with urination, with the exception of a weak urine stream and hesitancy, which he has had for the past 2 years. At the time of this visit, the patient is afebrile; the blood pressure is 130/88 mm Hg and the heart rate is 80/min. There is no flank tenderness. A urinalysis reveals no leukocytes and is negative for esterase. The urinalysis reveals 2 red blood cells (RBCs)/ high power field (HPF), and there are no casts on urinary sediment analysis. The physician, however, notices the following abnormality:
Prior treatment
BUN 12 mg/dL
Creatinine 1.2 mg/dL
Today’s visit
BUN 13 mg/dL
Creatinine 2.1 mg/dL \\

\noindent A: Let's think step by step. The patient has a history of hypertension and benign prostate hyperplasia. These are both risk factors for urinary tract infections. The patient is being treated with trimethoprim-sulfamethoxazole for a urinary tract infection. The symptoms have resolved, but the patient has a weak urine stream and hesitancy. The urinalysis reveals 2 red blood cells (RBCs)/ high power field (HPF), and there are no casts on urinary sediment analysis. The physician, however, notices the following abnormality: BUN 12 mg/dL, Creatinine 1.2 mg/dL (prior treatment) vs BUN 13 mg/dL, Creatinine 2.1 mg/dL (today’s visit). The BUN and creatinine levels have increased from the prior visit. This could be due to acute interstitial nephritis. A known but rare side effect of trimethoprim-sulfamethoxazole. Interstitial nephritis should be suspected if there is an elevated creatinine and a urinalysis that shows white cells, white cell casts, and, in some cases, eosinophiluria. Since none of the urinalysis results is present and the patient has no typical symptoms such as fever, rash, or acute interstitial nephritis is an unlikely diagnosis.
Answer: Reassure the patient, stop trimethoprim-sulfamethoxazole and repeat the measurement in 1–2 weeks. \\

\noindent Q: A 32-year-old woman comes to the physician because of weight gain, generalized weakness, and irregular menstrual cycles for the past 16 months. She began having symptoms of insomnia and depression 10 months ago. More recently, she has been having difficulties rising from a chair. She has a 2-year history of hypertension. Current medications include citalopram and hydrochlorothiazide. She is 168 cm (5 ft 6 in) tall and weighs 100 kg (220 lb). BMI is 36 kg/m2. Her pulse is 92/min, respirations are 18/min, and blood pressure is 134/76 mm Hg. She appears tired and has a full, plethoric face and central obesity. Examination of the skin shows violaceous linear striations on her lower abdomen. Two midnight serum cortisol studies show measurements of 288 \textmu g/L and 253 \textmu g/L (N < 90); a 24-hour urinary cortisol measurement was 395 \textmu g (N < 300). Upon follow-up laboratory examination, the patient's serum ACTH levels were also elevated at 136 pg/mL (N = 7–50). What is the most appropriate next step in evaluation? \\

\noindent A: Let's think step by step. The patient has Cushing's syndrome, which is characterized by weight gain, generalized weakness, and irregular menstrual cycles. She also has symptoms of insomnia and depression, which could be caused by Cushing's syndrome. Additionally, she has difficulties rising from a chair, which could be caused by muscle weakness. The two elevated serum cortisol measurements and the elevated ACTH levels confirm the diagnosis of Cushing's syndrome. A high dose dexamethasone suppression test will help determine if the cause of Cushing's syndrome is Cushing's disease (CS caused by pituitary hypersecretion of ACTH) or ectopic ACTH syndrome (CS caused by nonpituitary ACTH-secreting tumors). Since the diagnosis of Cushing's syndrome is established, the most appropriate next step is to distinguish between Cushing's disease and ectopic ACTH syndrome using the high-dose dexamethasone suppression test.
Answer: High-dose dexamethasone suppression test. \\
 
\noindent Q: A 53-year-old man comes to the physician because of a 3-month history of a nonpruritic rash. He has been feeling more tired than usual and occasionally experiences pain in his wrists and ankles. He does not smoke or drink alcohol. His temperature is 37.6°C (99.7°F), pulse is 98/min, respirations are 18/min, and blood pressure is 130/75 mm Hg. Physical examination shows multiple, erythematous, purpuric papules on his trunk and extremities that do not blanch when pressed. The remainder of the examination shows no abnormalities. The patient's hemoglobin is 14 g/dL, leukocyte count is 9,500/mm3, and platelet count is 228,000/mm3. Urinalysis and liver function tests are within normal limits. The test for rheumatoid factor is positive. Serum ANA is negative. Serum complement levels are decreased. Serum protein electrophoresis and immunofixation shows increased gammaglobulins with pronounced polyclonal IgM and IgG bands. Testing for cryoglobulins shows no precipitate after 24 hours. Chest x-ray and ECG show no abnormalities. What is the most appropriate next step in management? \\

\noindent A: Let's think step by step. The patient has a rash, fatigue, and pain in his wrists and ankles. He has normal hemoglobin (normal range 13.8-17.2 g/dL), normal leukocyte count (normal range 4500-11000 wbc/microliter), and normal platelet count (normal range 150-450 platelets/microliter). His serum complement levels are decreased. His serum protein electrophoresis and immunofixation show increased gammaglobulins with pronounced polyclonal IgM and IgG bands. The cryoglobulin precipitation test is normal, ruling out cryoglobulinemia. Serum ANA is negative, so Lupus is less likely. Physical examination and symptoms are consistent with possible non-cryoglobulinemic vasculitis. The most likely associated diagnosis is Hepatitis C.
Answer: Hepatitis C serology. \\

\section{Appendix 2: Datasets released}
\allowdisplaybreaks
\noindent

\noindent This appendix provides examples of questions from the datasets released, \MedQAOriginal{}, \MedQAOpen{} and \ClinicianCases{}. We also demonstrate the conversion of an example from \MedQAOriginal{} to \MedQAOpen{} format.

\subsection{MedQA Dataset}
\begin{enumerate}
    \item \textbf{\MedQAOriginal{} Question No. 1}
    
    \textit{A 30-year-old G4P3 woman at 38 weeks gestation is admitted to the labor and delivery unit complaining of contractions every 5 minutes for the past hour. Her previous births have been via uncomplicated caesarean sections, but she wishes to attempt vaginal delivery this time. Her prenatal care is notable for gestational diabetes controlled with diet and exercise. The delivery is prolonged, but the patient's pain is controlled with epidural analgesia. She delivers a male infant with Apgar scores of 8 and 9 at 1 and 5 minutes, respectively. Fundal massage is performed, but the placenta does not pass. The obstetrician manually removes the placenta, but a red mass protrudes through the vagina attached to the placenta. The patient loses 500 mL of blood over the next minute, during which her blood pressure decreases from 120/80 mmHg to 90/65 mmHg. What is the best next step in management?} \\
    
    \noindent\textbf{Options: } {'A': 'Hysterectomy', 'B': 'Intravenous oxytocin', 'C': 'Elevate posterior fornix', 'D': 'Red blood cell transfusion'} \\
    
    \noindent\textbf{Correct Answer: } The correct answer is \textbf{C: Elevate posterior fornix} 
    
    \begin{itemize}
        \item \textbf{Converted Descriptive Version of the Question}
        A 30-year-old G4P3 woman at 38 weeks gestation is admitted to the labor and delivery unit complaining of contractions every 5 minutes for the past hour. Her previous births have been via uncomplicated caesarean sections, but she wishes to attempt vaginal delivery this time. Her prenatal care is notable for gestational diabetes controlled with diet and exercise. The delivery is prolonged, but the patient's pain is controlled with epidural analgesia. She delivers a male infant with Apgar scores of 8 and 9 at 1 and 5 minutes, respectively. Fundal massage is performed, but the placenta does not pass. The obstetrician manually removes the placenta, but a red mass protrudes through the vagina attached to the placenta. The patient loses 500 mL of blood over the next minute, during which her blood pressure decreases from 120/80 mmHg to 90/65 mmHg. Elaborate the steps of management of this case ?
           
        \item \textbf{Reasoning for the open question format}
        In this scenario, the patient has a retained placenta with a red mass protruding through the vagina, which suggests that there may be a placental abnormality such as a placenta accreta or placenta percreta. This can cause excessive bleeding, which can lead to hypovolemic shock and potentially life-threatening complications. The first step in managing this case would be to elevate the posterior fornix, which will help to reduce the bleeding by increasing the intrauterine pressure and compressing the bleeding vessels. This will also help to prevent further descent of the placenta and allow for easier removal. Additionally, the patient's blood loss should be closely monitored and fluid replacement should be administered as needed to maintain blood pressure and prevent shock. Thus, the answer is Elevate posterior fornix.
        
    \end{itemize}
    
    \item \textbf{\MedQAOriginal{} Question No. 2}
    
    \textit{A 16-year-old boy is brought to the emergency department by ambulance from a soccer game. During the game, he was about to kick the ball when another player collided with his leg from the front. He was unable to stand up after this collision and reported severe knee pain. On presentation, he was found to have a mild knee effusion. Physical exam showed that his knee could be pushed posteriorly at 90 degrees of flexion but it could not be pulled anteriorly in the same position. The anatomic structure that was most likely injured in this patient has which of the following characteristics?}
    
    \noindent\textbf{Options: } {'A': 'Runs anteriorly from the medial femoral condyle', 'B': 'Runs medially from the lateral femoral condyle', 'C': 'Runs posteriorly from the lateral femoral condyle', 'D': 'Runs posteriorly from the medial femoral condyle'} \\
    
    \noindent\textbf{Correct Answer: } The correct answer is \textbf{D: Runs posteriorly from the medial femoral condyle}
    
    \begin{itemize}
        \item \textbf{Converted Descriptive Version of the Question}
        A 16-year-old boy is brought to the emergency department by ambulance from a soccer game. During the game, he was about to kick the ball when another player collided with his leg from the front. He was unable to stand up after this collision and reported severe knee pain. On presentation, he was found to have a mild knee effusion. Physical exam showed that his knee could be pushed posteriorly at 90 degrees of flexion but it could not be pulled anteriorly in the same position. Corresponding to the signs and symptoms, the patient presents with and the investigations performed already, what immediate measures of management are to be initiated, prioritising their urgency, and keeping in mind the probable diagnosis for the same?
        \item \textbf{Reasoning for the open question}
        The patient's history of sudden onset knee pain following a direct blow to the knee, along with the physical examination findings of a mild knee effusion and inability to pull the knee anteriorly at 90 degrees of flexion, suggests a possible diagnosis of a medial tibial stress syndrome (MTSS) or shin splints. This condition is caused by repetitive trauma or overuse of the tibia, leading to inflammation and pain in the surrounding soft tissues. The patient's age and participation in sports also increase the likelihood of this diagnosis. To confirm the diagnosis, imaging studies such as X-rays or an MRI may be ordered. In the meantime, conservative management options such as rest, ice, compression, and elevation (RICE) should be initiated to alleviate pain and reduce inflammation. Additionally, the patient should be advised to avoid activities that exacerbate the condition and consider wearing proper footwear and orthotics to prevent further injury. Thus, the answer is Runs posteriorly from the medial femoral condyle.
    \end{itemize}
  
    \item \textbf{\MedQAOriginal{} Question No. 3}
    
    \textit{A 64-year-old male with a past medical history of two myocardial infarctions presents to the emergency room with shortness of breath. He notes that he stopped taking his furosemide two weeks prior, because he ran out of pills. On exam, his oxygen saturation is 78\%, his lungs have crackles throughout, and jugular venous pulsation is located at the earlobe. EKG and troponin levels are normal. Which of the following is consistent with this man's pulmonary physiology?}
    
    \noindent\textbf{Options: } {'A': 'Decreased Aa gradient, decreased surface area for diffusion, normal diffusion distance', 'B': 'Decreased Aa gradient, increased surface area for diffusion, decreased diffusion distance', 'C': 'Increased Aa gradient, normal surface area for diffusion, increased diffusion distance', 'D': 'Increased Aa gradient, decreased surface area for diffusion, increased diffusion distance'} \\
    
    \noindent\textbf{Correct Answer: } The correct answer is \textbf{D: Increased Aa gradient, decreased surface area for diffusion, increased diffusion distance} 
    
    \begin{itemize}
        \item \textbf{Converted Descriptive Version of the Question}
        A 64-year-old male with a past medical history of two myocardial infarctions presents to the emergency room with shortness of breath. He notes that he stopped taking his furosemide two weeks prior, because he ran out of pills. On exam, his oxygen saturation is 78\%, his lungs have crackles throughout, and jugular venous pulsation is located at the earlobe. EKG and troponin levels are normal. What pulmonary pathology is this man suffering from?
        \item \textbf{Reasoning for the open question}
        The patient's symptoms of shortness of breath, low oxygen saturation, and lung crackles suggest respiratory distress. The absence of abnormal EKG and troponin levels rules out acute coronary syndrome or cardiac failure as the cause of his symptoms. The patient's history of stopping furosemide two weeks ago suggests that he may have developed fluid overload, which can lead to pulmonary edema. This is supported by the presence of jugular venous pulsation at the earlobe, indicating elevated central venous pressure. The patient's age and history of myocardial infarctions also increase his risk for developing pulmonary disease. Therefore, the most likely diagnosis is increased Aa gradient, decreased surface area for diffusion, and increased diffusion distance, consistent with pulmonary edema. Thus, the answer is Increased Aa gradient, decreased surface area for diffusion, increased diffusion distance.
    \end{itemize}
    
\end{enumerate}

\subsection{\ClinicianCases{} Dataset}

\begin{enumerate}
    \item \textbf{\ClinicianCases{} Question No. 1}
    
    \textit{A 44 year old woman visited the opd c/o heavy menstrual bleeding for the past 10 days preceded by prolonged spotting for 8 days. she has recently noticed some irregularity in her menses in the last 3-6 months. She has noticed weight gain of about 3 kg in the same time period. She has had two normal deliveries and no other surgeries. She has diabetes mellitus and has been on tab. Metformin 500 mg three times a day for the last 2 years. She reports that her blood sugars are regularly checked but recently have been outside the limit advised by her GP. She is not a known hypertensive. An ultrasonography of the pelvis reveals a bulky uterus 12*7*8 cm in size with an endometrial thickness of 18mm showing cystic spaces within. What should be the next investigation and treatment plan in this patient?} \\
    
    \noindent\textbf{Correct Answer: }
        The correct answer is \textbf{Pipelle endometrial biopsy followed by cyclical progesterone therapy or Mirena insertion,  dilatation and curettage for endometrial biopsy or  hysteroscopy with endometrial biopsy , Followed by cyclical progesterone or Mirena insertion}
    \begin{itemize}
        \item \textbf{Reasoning for the no-opt question}
          The patient's symptoms of heavy menstrual bleeding and irregular periods suggest endometrial pathology. Her recent weight gain and history of diabetes increase the likelihood of endometrial cancer. The ultrasound findings of a bulky uterus with cystic spaces within the endometrium further support this diagnosis. To confirm the diagnosis, a pipelle endometrial biopsy should be performed. If the biopsy results indicate endometrial cancer, treatment options include cyclical progesterone therapy or Mirena insertion, or dilatation and curettage for endometrial biopsy or hysteroscopy with endometrial biopsy. These treatments aim to reduce the risk of recurrent bleeding and prevent further complications. Thus, the answer is Pipelle endometrial biopsy followed by cyclical progesterone therapy or Mirena insertion, or dilatation and curettage for endometrial biopsy or hysteroscopy with endometrial biopsy, followed by cyclical progesterone or Mirena insertion.
        
    \end{itemize}
    
    \item \textbf{\ClinicianCases{} Question No. 2}
    
    \textit{A 27 year old man presented with red painful nodules over both shins for five days. He also gives a history of persistent dry cough and shortness of breath. On further enquiry, he experiences blurred vision and occasional pain in his eyes. On examination, there was a single palpable asymptomatic nodule in the right axilla. Which one blood test will confirm the diagnosis?} \\
    
    \noindent\textbf{Correct Answer: }
        The correct answer is \textbf{ACE levels} 
    \begin{itemize}
        \item \textbf{Reasoning for the no-opt question}
         The patient's symptoms of red painful nodules on both shins, persistent dry cough, shortness of breath, blurred vision, and occasional eye pain are consistent with a diagnosis of sarcoidosis, a systemic inflammatory disease caused by the growth of granulomas. Elevated levels of angiotensin-converting enzyme (ACE) in the blood are commonly seen in patients with sarcoidosis, and can be used to support the diagnosis. Therefore, measuring ACE levels in this patient's blood would help confirm the diagnosis of sarcoidosis. Thus, the answer is ACE levels.
    \end{itemize}
  
    \item \textbf{\ClinicianCases{} Question No. 3}
    
    \textit{A 46 year old man visits an outpatient department of dermatology with sudden loss of scalp hair in patches. On a detailed examination, partial loss of eyelashes were noticed. Hair pull test was positive at the margins of the bald patches. On dermoscopic evaluation, black dots, yellow dots, broken and tapering hair were observed. What is the drug of choice to halt the progression of the disease?} \\
    
    \noindent\textbf{Correct Answer}
        The correct answer is \textbf{Corticosteroids}
    \begin{itemize} 
        \item \textbf{Reasoning for the no-opt question}
         Alopecia areata is an autoimmune condition that causes hair loss in patches or completely. In this case, the patient has partial loss of eyelashes as well, which supports the diagnosis. The presence of black dots, yellow dots, broken and tapering hair on dermoscopy suggests that there is an active inflammation in the area. Corticosteroids are effective in halting the progression of alopecia areata by reducing inflammation and suppressing the immune system. Topical corticosteroids can be applied directly to the affected areas, while intralesional corticosteroid injections can be used for more resistant cases. Thus, the answer is corticosteroids.
    \end{itemize}
    
\end{enumerate}
\section{Appendix 3: Examples of Chosen/Rejected Reasoning Pairs for Reward Model training}
s
\allowdisplaybreaks
\noindent

\subsection{Examples of Chosen/Rejected Reasoning Pairs for Reward Model training}
In this section, we discuss the need to create chosen and rejected reasoning pairs for training the verifier for our model. \\
The chosen reasoning is the reasoning that led to the correct answer, while the rejected reasoning is the reasoning that led to incorrect answers. By training the verifier with both chosen and rejected reasoning pairs, the model can learn to distinguish between valid and invalid reasoning processes, improving its accuracy in answering questions. \\

\subsection{Example Question No 1}

\textbf{Question:} A 67-year-old woman presents to her primary care physician for urinary incontinence. She has been waking up every morning with her underwear soaked in urine. She notices that at work if she does not take regular bathroom breaks her underwear will have some urine in it. She urinates 5 to 11 times per day but she claims it is a small volume. Her current medications include lisinopril, metformin, insulin, aspirin, atorvastatin, sodium docusate, and loratadine. Her temperature is 98.2°F (36.8°C), blood pressure is 167/108 mmHg, pulse is 90/min, respirations are 15/min, and oxygen saturation is 99\% on room air. Physical exam is notable for decreased pinprick sensation in the lower extremities and a systolic murmur along the right upper sternal border. Which of the following is the best treatment for this patient? \\

\noindent\textbf{Options :}
{'A': 'Bethanechol and intermittent straight catheterization', 'B': 'Bethanechol during the day and oxybutynin at night', 'C': 'No recommendations needed', 'D': 'Reduce fluid intake and discontinue diuretics'} \\

\noindent\textbf{Correct Answer: }
The correct answer is \textbf{Bethanechol and intermittent straight catheterization}.

\subsubsection{Chosen and Rejected Reasoning Pairs}

\noindent\textbf{Pair 1:} \\

\noindent\textit{Chosen Reasoning:}
    The patient's symptoms of urinary incontinence, frequent urination, and decreased pinprick sensation in the lower extremities suggest a neurological cause, such as a spinal cord injury or multiple sclerosis. However, the patient's history of hypertension, hyperglycemia, and hyperlipidemia also suggest a possible contribution from an overactive bladder. Therefore, the best treatment approach would be to start the patient on bethanechol, a cholinergic agent that can help improve bladder function and reduce urinary frequency. Additionally, intermittent straight catheterization can help manage urinary retention and prevent urinary tract infections. This combination of medication and catheterization can effectively address both the neurological and non-neurological contributors to the patient's urinary incontinence, thus improving her quality of life. Thus, the answer is Bethanechol and intermittent straight catheterization. \\
    
\noindent\textit{Rejected Reasoning:}
    The patient's symptoms of urinary incontinence, frequent urination, and decreased pinprick sensation in the lower extremities suggest a neurological cause, such as overactive bladder or spinal cord injury. The patient's history of hypertension and diabetes mellitus also increase the likelihood of a neurological cause. Bethanechol, a cholinergic agonist, is used to treat urinary retention and can help improve bladder function by increasing the frequency and amplitude of contractions. Oxybutynin, an anticholinergic agent, can help reduce urgency and frequency of urination, especially at night when the patient may not be able to access a bathroom quickly. Thus, the combination of bethanechol during the day and oxybutynin at night would be the best treatment approach for this patient. \\
    
\noindent\textbf{Pair 2:} \\

\noindent\textit{Chosen Reasoning:}
 The patient's symptoms of urinary incontinence, frequent urination, and decreased pinprick sensation in the lower extremities suggest a neurological cause, such as a spinal cord injury or multiple sclerosis. However, the patient's history of hypertension, hyperglycemia, and hyperlipidemia also suggest a possible contribution from an overactive bladder. Therefore, the best treatment approach would be to start the patient on bethanechol, a cholinergic agent that can help improve bladder function and reduce urinary frequency. Additionally, intermittent straight catheterization can help manage urinary retention and prevent urinary tract infections. This combination of medication and catheterization can effectively address both the neurological and non-neurological contributors to the patient's urinary incontinence, thus improving her quality of life. Thus, the answer is Bethanechol and intermittent straight catheterization. \\

\noindent\textit{Rejected Reasoning:}
 This patient's symptoms of urinary incontinence, frequency, and nocturia are consistent with overactive bladder (OAB). However, her age, hypertension, diabetes mellitus, and neuropathy suggest that she may also have underlying detrusor instability or urgency urinary incontinence. Given these factors, the best course of action would be to refer her to a urologist for further evaluation and management. A urologist can perform tests such as a urinalysis, postvoid residual measurement, and cystoscopy to confirm the diagnosis and determine the appropriate treatment. Thus, the answer is "No recommendations needed" as the patient requires specialized care beyond the scope of primary care. \\

\noindent\textbf{Pair 3:} \\

\noindent\textit{Chosen Reasoning:}
 The patient's symptoms of urinary incontinence, frequent urination, and decreased pinprick sensation in the lower extremities suggest a neurological cause, such as a spinal cord injury or multiple sclerosis. However, the patient's history of hypertension, hyperglycemia, and hyperlipidemia also suggest a possible contribution from an overactive bladder. Therefore, the best treatment approach would be to start the patient on bethanechol, a cholinergic agent that can help improve bladder function and reduce urinary frequency. Additionally, intermittent straight catheterization can help manage urinary retention and prevent urinary tract infections. This combination of medication and catheterization can effectively address both the neurological and non-neurological contributors to the patient's urinary incontinence, thus improving her quality of life. Thus, the answer is Bethanechol and intermittent straight catheterization. \\

\noindent\textit{Rejected Reasoning:}
 The patient's symptoms of urinary incontinence and frequent urination suggest an overactive bladder. This can be caused by excessive fluid intake or certain medications such as diuretics. In this case, the patient's history of taking lisinopril, a diuretic, likely contributes to her symptoms. Additionally, her high blood pressure and systolic murmur suggest that reducing fluid intake may help alleviate these issues. Discontinuing the diuretic medication and limiting fluid intake would help reduce the frequency of urination and minimize the risk of incontinence. Thus, the answer is Reduce fluid intake and discontinue diuretics. \\

\subsection{Example Question No 2}

\textbf{Question:} A 55-year-old man presents into the emergency department with a severe cough and difficulty breathing. He says that he finds himself out of breath after taking a few steps, and has to sit down and rest, in order to continue. He also says that, at night, he has the greatest difficulty in breathing and usually uses at least 3 pillows to sleep comfortably. He mentions a cough that appears only at night, but which is persistent enough to wake him up from sleep. He mentions that he has had a ‘heart attack’ 5 years ago. He also says that he continues to consume alcohol on a regular basis even though his doctor has advised against it. He has brought his lab reports which he had recently got done on the suggestions of his family doctor. An electrocardiogram (ECG) and a chest X-ray are found. Which of the following is the next step in this patient’s management? \\

\noindent\textbf{Options: }
{'A': 'Computed Tomography (CT)', 'B': 'Echocardiogram', 'C': 'Stress test', 'D': 'Cardiac biopsy'} \\

\noindent\textbf{Correct Answer: }
The correct answer is \textbf{Echocardiogram}. 

\subsubsection{Chosen and Rejected Reasoning Pairs}

\noindent\textbf{Pair 1:} \\

\noindent\textit{Chosen Reasoning:}
 This patient's symptoms of shortness of breath, especially at night, and use of multiple pillows to sleep suggest pulmonary congestion. His history of heart attack increases the likelihood of underlying cardiac disease. The presence of a persistent nighttime cough further supports this diagnosis. Additionally, the patient's continued alcohol consumption may have contributed to the development or exacerbation of any cardiac condition. An echocardiogram would provide valuable information regarding the structure and function of the heart, allowing for the assessment of potential cardiac causes of his symptoms, such as left ventricular dysfunction or valvular heart disease. Thus, the answer is Echocardiogram. \\

\noindent\textit{Rejected Reasoning:}
 This patient's symptoms of shortness of breath, particularly at night, and use of multiple pillows to sleep suggest possible pulmonary disease. His history of heart attack increases the likelihood of coronary artery disease. Alcohol consumption further increases the risk of lung disease. A chest X-ray does not provide adequate information to diagnose pulmonary embolism or lung cancer, which are possibilities based on the patient's symptoms and history. An ECG may help diagnose any cardiac arrhythmias, but it will not provide information about lung pathology. A CT scan of the lungs is the most appropriate next step in this patient's management as it can help diagnose pulmonary embolism, lung cancer, and other lung diseases. Thus, the answer is Computed Tomography (CT). \\

\noindent\textbf{Pair 2:} \\

\noindent\textit{Chosen Reasoning:}
 This patient's symptoms of shortness of breath, especially at night, and use of multiple pillows to sleep suggest pulmonary congestion. His history of heart attack increases the likelihood of underlying cardiac disease. The presence of a persistent nighttime cough further supports this diagnosis. Additionally, the patient's continued alcohol consumption may have contributed to the development or exacerbation of any cardiac condition. An echocardiogram would provide valuable information regarding the structure and function of the heart, allowing for the assessment of potential cardiac causes of his symptoms, such as left ventricular dysfunction or valvular heart disease. Thus, the answer is Echocardiogram. \\

\noindent\textit{Rejected Reasoning:}
 This patient's symptoms of shortness of breath, especially at night, and a history of heart attack suggest that he may have coronary artery disease or other cardiac issues. The fact that he experiences difficulty breathing when lying flat and requires multiple pillows to sleep suggests that he may have pulmonary congestion or left ventricular failure. Additionally, his persistent cough at night could be indicative of acid reflux or aspiration. Given these concerns, a stress test would be an appropriate next step in the patient's management to assess his cardiac function and identify any potential issues. Thus, the answer is Stress test. \\

\noindent\textbf{Pair 3:} \\

\noindent\textit{Chosen Reasoning:}
 This patient's symptoms of shortness of breath, especially at night, and use of multiple pillows to sleep suggest pulmonary congestion. His history of heart attack increases the likelihood of underlying cardiac disease. The presence of a persistent nighttime cough further supports this diagnosis. Additionally, the patient's continued alcohol consumption may have contributed to the development or exacerbation of any cardiac condition. An echocardiogram would provide valuable information regarding the structure and function of the heart, allowing for the assessment of potential cardiac causes of his symptoms, such as left ventricular dysfunction or valvular heart disease. Thus, the answer is Echocardiogram. \\

\noindent\textit{Rejected Reasoning:}
 The patient's symptoms of shortness of breath, especially at night, and persistent cough suggest chronic obstructive pulmonary disease (COPD). His history of heart attack and continued alcohol consumption increase the likelihood of cardiac involvement. The ECG and chest X-ray do not provide sufficient information to diagnose COPD or assess cardiac function. A cardiac biopsy would help establish the diagnosis of COPD and determine the extent of cardiac damage, guiding appropriate management and treatment. Thus, the answer is Cardiac biopsy. \\

\section{Appendix 4: Examples of Responses generated using \OpenClinicR{} and \OpenEliminative{} approaches}
s\allowdisplaybreaks
\noindent

\noindent This section provides more examples of the Responses generated using \OpenClinicR{} and \OpenEliminative{} approaches highlighting their differences. \\

\subsection{Question No. 1}

\noindent \textbf{Question: }

\noindent A male newborn is evaluated 24 hours after delivery for high-pitched crying, poor feeding, rhinorrhea, and low-grade fever. He was born at 40 weeks' gestation at 2514 g (5.54 lb) to a 28-year-old woman, gravida 3, para 2, by an uncomplicated cesarean section. Apgar scores were 8 and 9 at 1 and 5 minutes, respectively. The mother did not receive prenatal care. The infant's temperature is 38.0Â°C (100.4Â°F), pulse is 170/min, and blood pressure is 71/39 mm Hg. Examination shows hyperreflexia, tremors, and an excessive startle response.  These symptoms are mostly like due to maternal use of what substance? \\

\noindent \textbf{Response generated using \OpenClinicR{} approach: }

\noindent The baby is showing symptoms such as high pitched crying, poor feeding, rhinorrhea and low grade fever. The baby was born full term at 40 weeks and had Apgar scores of 8 and 9 at 1 and 5 minutes which indicates that there was no distress during delivery. Mother did not receive prenatal care which could indicate lack of awareness or lack of resources. Baby has hyperreflexia, tremors and excessive startle response which could indicate neurological problems. It could be due to maternal drug use during pregnancy. Maternal drug use during pregnancy can cause withdrawal symptoms in the baby after birth. The most common drugs that cause withdrawal symptoms in the baby are opioids, benzodiazepines, alcohol, nicotine, and cocaine. As the mother did not receive prenatal care, it is more likely that the mother used drugs during pregnancy. Therefore, the baby's symptoms are most likely due to maternal use of drugs during pregnancy.
Answer: maternal use of drugs during pregnancy. \\

\noindent \textbf{Response generated using \OpenEliminative{} approach: }

\noindent The baby has a high-pitched cry, poor feeding, rhinorrhea, and low-grade fever. The baby was born at 40 weeks' gestation at 2514 g (5.54 lb) to a 28-year-old woman, gravida 3, para 2, by an uncomplicated cesarean section. Apgar scores were 8 and 9 at 1 and 5 minutes, respectively. The mother did not receive prenatal care. The baby's temperature is 38.0Â°C (100.4Â°F), pulse is 170/min, and blood pressure is 71/39 mm Hg. Examination shows hyperreflexia, tremors, and an excessive startle response. These symptoms are most likely due to withdrawal from maternal drug use during pregnancy. The most common drugs that cause withdrawal symptoms in neonates are opioids, benzodiazepines, and stimulants. The mother did not receive prenatal care, which suggests that she may have used drugs during pregnancy. The baby's symptoms are consistent with withdrawal from opioids or benzodiazepines.
 Answer: Opioids or Benzodiazepines. \\

\subsection{Question No. 2}

\noindent \textbf{Question: }

\noindent A 59-year-old man presents to general medical clinic for his yearly checkup. He has no complaints except for a dry cough. He has a past medical history of type II diabetes, hypertension, hyperlipidemia, asthma, and depression. His home medications are sitagliptin/metformin, lisinopril, atorvastatin, albuterol inhaler, and citalopram. His vitals signs are stable, with blood pressure 126/79 mmHg. Hemoglobin A1C is 6.3\%, and creatinine is 1.3 g/dL. The remainder of his physical exam is unremarkable. If this patient's cough is due to one of the medications he is taking, what would be the next step in management? \\

\noindent \textbf{Response generated using \OpenClinicR{} approach: }

\noindent The patient has a history of multiple comorbidities such as diabetes, hypertension, hyperlipidemia, asthma, and depression. He is on several medications including sitagliptin/metformin, lisinopril, atorvastatin, albuterol inhaler, and citalopram. The patient presents with a dry cough, which could be related to one of the medications he is taking. The cough could be due to the ACE inhibitor lisinopril, which is a known side effect. The next step in management would be to hold the lisinopril and monitor the patient's blood pressure to ensure it remains controlled. Additionally, an alternative medication for hypertension could be considered if the patient's blood pressure remains stable.
Answer: Hold lisinopril and monitor blood pressure. \\

\noindent \textbf{Response generated using \OpenEliminative{} approach: }

\noindent The patient has a dry cough. It could be related to his medications. The patient takes several medications, including sitagliptin/metformin, lisinopril, atorvastatin, albuterol inhaler, and citalopram. The cough could be caused by any of these medications. However, it is important to note that a dry cough is a common side effect of ACE inhibitors, such as lisinopril. Therefore, the next step in management would be to consider alternative medications for hypertension, such as beta-blockers, calcium channel blockers, or thiazide diuretics.
Answer: Consider alternative medications for hypertension. \\
\section{Appendix 5: Instructions provided to the interns for evaluation of the experiments' results}
\allowdisplaybreaks

\noindent This section gives the instructions that we provided to the interns before the result sheets were provided to them for evaluation. \\

\subsection{Instructions}
\noindent \texttt{Each row in the task sheet contains a medical question (Column A) and reasoning generated by different approaches in a shuffled order (Columns B, D, F, H) in the excel sheet provided. The task is to evaluate the reasoning based on its quality on a 3-point scale described below:} \\

\begin{itemize}
    \item \texttt{Agree: the answer is perfect with no mistakes OR the answer is mostly correct, with one or two minor mistakes} 
    \item \texttt{Neutral: the answer is not incorrect, but there is missing information OR it is hard to decide given my knowledge} 
    \item \texttt{Disagree: the answer is incorrect but it has a few things right OR the answer is completely hallucinated} 
\end{itemize}

\noindent \texttt{Each Response column will be followed by a drop-down list column containing the options for the 3-point scale. Please select one from the dropdown list.} \\

\noindent \texttt{Please feel free to reach out to me in case of any doubt OR if you have any queries.} \\

\newpage
\section{Medical Expert Evaluation results of all the experiments }

\noindent

\subsection{Results of Human Evaluations of Llama-2-7B chat responses on \ClinicianCases{} dataset}
\begin{table}[h!]
    \centering
    \begin{tabular}{l|c|c|c}
    \hline
       & Agree & Neutral & Disagree\\
       \hline
       \OpenClinicR{}& 88 & 4 & 8 \\
       \OpenEliminative{}& 92 & 0 & 8 \\
       \OpenClinicR{}-\MCQEliminative{}& 92 & 8 & 0 \\
       \OpenClinicR{}-Verifier & 92 & 4 & 4 \\
    \end{tabular}
    \caption{Medical Expert-1 evaluated results of all four approaches on \ClinicianCases{} dataset (25 questions)}
    \label{tab:my_label}
\end{table}

\begin{table}[h!]
    \centering
    \begin{tabular}{l|c|c|c}
    \hline
       & Agree & Neutral & Disagree\\
       \hline
       \OpenClinicR{}& 88 & 8 & 4 \\
       \OpenEliminative{}& 88 & 4 & 8 \\
       \OpenClinicR{}-\MCQEliminative{}& 84 & 12 & 4 \\
       \OpenClinicR{}-Verifier & 88 & 4 & 8 \\
    \end{tabular}
    \caption{Medical Expert-2 evaluated results of all four approaches on \ClinicianCases{} dataset (25 questions)}
    \label{tab:my_label}
\end{table}

\noindent The average inter-annotator agreement for this section: \textbf{0.860} \\

\subsection{Results of Human Evaluations of Llama-2-7B chat responses on \MedQAOpen{} dataset}

\begin{table}[h!]
    \centering
    \begin{tabular}{l|c|c|c}
    \hline
       & Agree & Neutral & Disagree\\
       \hline
       \OpenClinicR{}& 94 & 4 & 2 \\
       \OpenEliminative{}& 49 & 13 & 38 \\
       \OpenClinicR{}-\MCQEliminative{}& 84 & 11 & 5 \\
       \OpenClinicR{}-Verifier & 95 & 3 & 2 \\
    \end{tabular}
    \caption{Medical Expert-1 evaluated results of all four approaches on \MedQAOpen{} dataset Set 1 (100 questions)}
    \label{tab:my_label}
\end{table}
\begin{table}[h!]
    \centering
    \begin{tabular}{l|c|c|c}
    \hline
       & Agree & Neutral & Disagree\\
       \hline
       \OpenClinicR{}& 94 & 4 & 2 \\
       \OpenEliminative{}& 54 & 16 & 30 \\
       \OpenClinicR{}-\MCQEliminative{}& 85 & 11 & 4 \\
       \OpenClinicR{}-Verifier & 96 & 2 & 2 \\
    \end{tabular}
    \caption{Medical Expert-2 evaluated results of all four approaches on \MedQAOpen{} dataset Set 1 (100 questions)}
    \label{tab:my_label}
\end{table}

\begin{table}[h!]
    \centering
    \begin{tabular}{l|c|c|c}
    \hline
       & Agree & Neutral & Disagree\\
       \hline
       \OpenClinicR{}& 88 & 0 & 12 \\
       \OpenEliminative{}& 54 & 7 & 39 \\
       \OpenClinicR{}-\MCQEliminative{}& 77 & 1 & 22 \\
       \OpenClinicR{}-Verifier & 85 & 1 & 14 \\
    \end{tabular}
    \caption{Medical Expert-1 evaluated results of all four approaches on \MedQAOpen{} dataset Set 2 (100 questions)}
    \label{tab:my_label}
\end{table}
\clearpage

\begin{table}[h!]
    \centering
    \begin{tabular}{l|c|c|c}
    \hline
       & Agree & Neutral & Disagree\\
       \hline
       \OpenClinicR{}& 89 & 0 & 11 \\
       \OpenEliminative{}& 55 & 6 & 39 \\
       \OpenClinicR{}-\MCQEliminative{}& 79 & 0 & 21 \\
       \OpenClinicR{}-Verifier & 86 & 0 & 14 \\
    \end{tabular}
    \caption{Medical Expert-2 evaluated results of all four approaches on \MedQAOpen{} dataset Set 2 (100 questions)}
    \label{tab:my_label}
\end{table}

\begin{table}[h!]
    \centering
    \begin{tabular}{l|c|c|c}
    \hline
       & Agree & Neutral & Disagree\\
       \hline
       \OpenClinicR{}& 65 & 31 & 4 \\
       \OpenEliminative{}& 68 & 27 & 5 \\
       \OpenClinicR{}-\MCQEliminative{}& 69 & 25 & 6 \\
       \OpenClinicR{}-Verifier & 76 & 21 & 3 \\
    \end{tabular}
    \caption{Medical Expert-1 evaluated results of all four approaches on \MedQAOpen{} dataset Set 3 (100 questions)}
    \label{tab:my_label}
\end{table}
\begin{table}[h!]
    \centering
    \begin{tabular}{l|c|c|c}
    \hline
       & Agree & Neutral & Disagree\\
       \hline
       \OpenClinicR{}& 92 & 2 & 6 \\
       \OpenEliminative{}& 62 & 10 & 28 \\
       \OpenClinicR{}-\MCQEliminative{}& 89 & 1 & 10 \\
       \OpenClinicR{}-Verifier & 95 & 1 & 4 \\
    \end{tabular}
    \caption{Medical Expert-2 evaluated results of all four approaches on \MedQAOpen{} dataset Set 3 (100 questions)}
    \label{tab:my_label}
\end{table}

\begin{table}[h!]
    \centering
    \begin{tabular}{l|c|c|c}
    \hline
       & Agree & Neutral & Disagree\\
       \hline
       \OpenClinicR{}& 70 & 19 & 11 \\
       \OpenEliminative{}& 13 & 32 & 55 \\
       \OpenClinicR{}-\MCQEliminative{}& 73 & 17 & 10 \\
       \OpenClinicR{}-Verifier & 76 & 15 & 9 \\
    \end{tabular}
    \caption{Medical Expert-1 evaluated results of all four approaches on \MedQAOpen{} dataset Set 4 (100 questions)}
    \label{tab:my_label}
\end{table}
\begin{table}[h!]
    \centering
    \begin{tabular}{l|c|c|c}
    \hline
       & Agree & Neutral & Disagree\\
       \hline
       \OpenClinicR{}& 70 & 16 & 14 \\
       \OpenEliminative{}& 46 & 16 & 38 \\
       \OpenClinicR{}-\MCQEliminative{}& 76 & 13 & 11 \\
       \OpenClinicR{}-Verifier & 81 & 13 & 6 \\
    \end{tabular}
    \caption{Medical Expert-2 evaluated results of all four approaches on \MedQAOpen{} dataset Set 4 (100 questions)}
    \label{tab:my_label}
\end{table}

\begin{table}[h!]
    \centering
    \begin{tabular}{l|c|c|c}
    \hline
       & Agree & Neutral & Disagree\\
       \hline
       \OpenClinicR{}& 84 & 1 & 15 \\
       \OpenEliminative{}& 78 & 1 & 21 \\
       \OpenClinicR{}-\MCQEliminative{}& 89 & 0 & 11 \\
       \OpenClinicR{}-Verifier & 91 & 0 & 9 \\
    \end{tabular}
    \caption{Medical Expert-1 evaluated results of all four approaches on \MedQAOpen{} dataset Set 5 (100 questions)}
    \label{tab:my_label}
\end{table}
\clearpage

\begin{table}[h!]
    \centering
    \begin{tabular}{l|c|c|c}
    \hline
       & Agree & Neutral & Disagree\\
       \hline
       \OpenClinicR{}& 87 & 7 & 6 \\
       \OpenEliminative{}& 82 & 10 & 8 \\
       \OpenClinicR{}-\MCQEliminative{}& 86 & 9 & 5 \\
       \OpenClinicR{}-Verifier & 87 & 7 & 6 \\
    \end{tabular}
    \caption{Medical Expert-2 evaluated results of all four approaches on \MedQAOpen{} dataset Set 5 (100 questions)}
    \label{tab:my_label}
\end{table}

\noindent The average inter-annotator agreement for this section: \textbf{0.32} \\

\subsection{Results of Human Evaluations of Llama-2-70B chat responses on \ClinicianCases{} dataset}
\begin{table}[h!]
    \centering
    \begin{tabular}{l|c|c|c}
    \hline
       & Agree & Neutral & Disagree\\
       \hline
       \OpenClinicR{}& 92 & 0 & 8 \\
       \OpenEliminative{}& 96 & 0 & 4 \\
       \OpenClinicR{}-\MCQEliminative{}& 92 & 0 & 8 \\
       \OpenClinicR{}-Verifier & 88 & 0 & 12 \\
    \end{tabular}
    \caption{Medical Expert-1 evaluated results of all four approaches on \ClinicianCases{} dataset (25 questions)}
    \label{tab:my_label}
\end{table}

\begin{table}[h!]
    \centering
    \begin{tabular}{l|c|c|c}
    \hline
       & Agree & Neutral & Disagree\\
       \hline
       \OpenClinicR{}& 72 & 16 & 12 \\
       \OpenEliminative{}& 76 & 8 & 16 \\
       \OpenClinicR{}-\MCQEliminative{}& 76 & 12 & 12 \\
       \OpenClinicR{}-Verifier & 76 & 8 & 16s \\
    \end{tabular}
    \caption{Medical Expert-2 evaluated results of all four approaches on \ClinicianCases{} dataset (25 questions)}
    \label{tab:my_label}
\end{table}

\noindent The average inter-annotator agreement for this section: \textbf{0.32} \\

\subsection{Results of Human Evaluations of Llama-2-70B chat responses on \MedQAOpen{} dataset}
\begin{table}[h!]
    \centering
    \begin{tabular}{l|c|c|c}
    \hline
       & Agree & Neutral & Disagree\\
       \hline
       \OpenClinicR{}& 94 & 1 & 5 \\
       \OpenEliminative{}& 94 & 0 & 6 \\
       \OpenClinicR{}-\MCQEliminative{}& 94 & 1 & 5 \\
       \OpenClinicR{}-Verifier & 97 & 1 & 2 \\
    \end{tabular}
    \caption{Medical Expert-1 evaluated results of all four approaches on \MedQAOpen{} dataset Set 1 (125 questions)}
    \label{tab:my_label}
\end{table}
\begin{table}[h!]
    \centering
    \begin{tabular}{l|c|c|c}
    \hline
       & Agree & Neutral & Disagree\\
       \hline
       \OpenClinicR{}& 86 & 9 & 5 \\
       \OpenEliminative{}& 90 & 6 & 4 \\
       \OpenClinicR{}-\MCQEliminative{}& 89 & 7 & 4 \\
       \OpenClinicR{}-Verifier & 85 & 10 & 5 \\
    \end{tabular}
    \caption{Medical Expert-2 evaluated results of all four approaches on \MedQAOpen{} dataset Set 2 (125 questions)}
    \label{tab:my_label}
\end{table}
\clearpage

\begin{table}[h!]
    \centering
    \begin{tabular}{l|c|c|c}
    \hline
       & Agree & Neutral & Disagree\\
       \hline
       \OpenClinicR{}& 95 & 4 & 1 \\
       \OpenEliminative{}& 90 & 6 & 5 \\
       \OpenClinicR{}-\MCQEliminative{}& 97 & 1 & 2 \\
       \OpenClinicR{}-Verifier & 90 & 6 & 4 \\
    \end{tabular}
    \caption{Medical Expert-1 evaluated results of all four approaches on \MedQAOpen{} dataset Set 2 (125 questions)}
    \label{tab:my_label}
\end{table}
\begin{table}[h!]
    \centering
    \begin{tabular}{l|c|c|c}
    \hline
       & Agree & Neutral & Disagree\\
       \hline
       \OpenClinicR{}& 95 & 4 & 1 \\
       \OpenEliminative{}& 90 & 4 & 6 \\
       \OpenClinicR{}-\MCQEliminative{}& 97 & 2 & 2 \\
       \OpenClinicR{}-Verifier & 90 & 6 & 5 \\
    \end{tabular}
    \caption{Medical Expert-2 evaluated results of all four approaches on \MedQAOpen{} dataset Set 2 (125 questions)}
    \label{tab:my_label}
\end{table}

\begin{table}[h!]
    \centering
    \begin{tabular}{l|c|c|c}
    \hline
       & Agree & Neutral & Disagree\\
       \hline
       \OpenClinicR{}& 87 & 2 & 11 \\
       \OpenEliminative{}& 90 & 0 & 10 \\
       \OpenClinicR{}-\MCQEliminative{}& 83 & 5 & 12 \\
       \OpenClinicR{}-Verifier & 86 & 3 & 10 \\
    \end{tabular}
    \caption{Medical Expert-1 evaluated results of all four approaches on \MedQAOpen{} dataset Set 3 (125 questions)}
    \label{tab:my_label}
\end{table}
\begin{table}[h!]
    \centering
    \begin{tabular}{l|c|c|c}
    \hline
       & Agree & Neutral & Disagree\\
       \hline
       \OpenClinicR{}& 59 & 23 & 18 \\
       \OpenEliminative{}& 61 & 28 & 11 \\
       \OpenClinicR{}-\MCQEliminative{}& 73 & 19 & 8 \\
       \OpenClinicR{}-Verifier & 95 & 1 & 4 \\
    \end{tabular}
    \caption{Medical Expert-2 evaluated results of all four approaches on \MedQAOpen{} dataset Set 3 (125 questions)}
    \label{tab:my_label}
\end{table}

\begin{table}[h!]
    \centering
    \begin{tabular}{l|c|c|c}
    \hline
       & Agree & Neutral & Disagree\\
       \hline
       \OpenClinicR{}& 94 & 0 & 6 \\
       \OpenEliminative{}& 82 & 0 & 18 \\
       \OpenClinicR{}-\MCQEliminative{}& 87 & 0 & 13 \\
       \OpenClinicR{}-Verifier & 94 & 0 & 6 \\
    \end{tabular}
    \caption{Medical Expert-1 evaluated results of all four approaches on \MedQAOpen{} dataset Set 4 (125 questions)}
    \label{tab:my_label}
\end{table}
\begin{table}[h!]
    \centering
    \begin{tabular}{l|c|c|c}
    \hline
       & Agree & Neutral & Disagree\\
       \hline
       \OpenClinicR{}& 84 & 10 & 6 \\
       \OpenEliminative{}& 76 & 13 & 11 \\
       \OpenClinicR{}-\MCQEliminative{}& 79 & 10 & 11 \\
       \OpenClinicR{}-Verifier & 80 & 13 & 7 \\
    \end{tabular}
    \caption{Medical Expert-2 evaluated results of all four approaches on \MedQAOpen{} dataset Set 4 (125 questions)}
    \label{tab:my_label}
\end{table}

\noindent The average inter-annotator agreement for this section: \textbf{0.372} \\

\end{document}